\documentclass[11pt,a4paper]{article}

\usepackage[utf8]{inputenc}
\usepackage[T1]{fontenc}
\usepackage{lmodern}
\usepackage[margin=1in]{geometry}
\usepackage{amsmath,amssymb,amsfonts}
\usepackage{graphicx}
\usepackage{booktabs}
\usepackage{tabularx}
\usepackage{multirow}
\usepackage{array}
\usepackage{caption}
\usepackage{setspace}
\usepackage[expansion=false]{microtype}
\usepackage{xcolor}
\usepackage[authoryear,round]{natbib}
\usepackage{authblk}
\usepackage{adjustbox}
\usepackage{xurl}
\usepackage{seqsplit}
\usepackage[colorlinks=true,breaklinks=true,linkcolor=blue!60!black,citecolor=blue!60!black,urlcolor=blue!60!black]{hyperref}

\newcommand{\code}[1]{\nolinkurl{#1}}
\providecommand{\doi}[1]{doi:\,\href{https://doi.org/#1}{\nolinkurl{#1}}}

\title{\textbf{Temporal Graph Learning for Forecasting Invoice-Network Liquidity Exposure: An Explainable Machine Learning Framework Using Real-World SME Receivables}}

\author[1]{Sai Srikanth Madugula\thanks{Corresponding Author. Email: saisrikanth.madugula\_phd.2023@woxsen.edu.in | \mbox{ORCID: \href{https://orcid.org/0000-0001-6479-8443}{0000-0001-6479-8443}}}}
\author[2]{Peplluis Esteva de la Rosa\thanks{Email: joseplluis.delarosa@udg.edu | \mbox{ORCID: \href{https://orcid.org/0000-0003-1412-4170}{0000-0003-1412-4170}}}}
\author[3]{Daya Shankar\thanks{Email: daya.shankar@woxsen.edu.in}}

\affil[1]{\small School of Technology, Woxsen University, Hyderabad, Telangana 502345, India}
\affil[2]{\small Universitat de Girona, 17004 Girona, Spain}
\affil[3]{\small School of Sciences, Woxsen University, Hyderabad, Telangana 502345, India}

\date{}

\begin{document}
\maketitle

\begin{abstract}
\noindent Building on prior structural analysis of temporal trade-credit networks \citep{madugula2026beyond}, this study tests whether cutoff-specific relational features and temporal graph embeddings improve out-of-sample forecasts of recorded invoice issuance and contractual maturity exposure. Throughout this manuscript, ``liquidity exposure'' refers strictly to recorded contractual face-value and maturity schedules rather than observed cash receipts, unpaid balances, default probability, or realized liquidity shortfall. The audited master ledger contains 749,952 analytical invoices issued during 2012--2023 across 75,343 distinct firms totaling EUR~99.705 billion in face value, excluding 1,240 terminal 2024 bridge records \citep{esteva2026dataset}. Monthly forecast origins yield 621,585 active-firm observations over 131 origins. We benchmark zero, naive, historical-rate, linear, ridge, random forest, histogram gradient boosting, XGBoost, and LightGBM tabular models against directed graph features, prior-complete-year static embeddings, and past-only orthogonally aligned rolling \texttt{node2vec} representations under identical XGBoost settings. Five held-out calendar years (2019--2023) are evaluated via horizon-purged expanding training and separate pre-test residual calibration. Because full-panel fitting exceeded available memory, each learned training fold uses at most 5,000 uniformly sampled eligible firm-origin rows while retaining all 606,641 held-out test rows. For the prespecified primary target---newly issued debtor-side face value due within 30 days---pooled $\log(1+y)$ mean absolute error (log-MAE) is 1.767 for rolling XGBoost versus 1.606 for tabular XGBoost. The equal-date temporal-minus-tabular contrast is 0.056 log units (95\% circular six-month block confidence interval $[-0.014, 0.132]$; Holm-adjusted $p = 0.262$), establishing no gain over tabular histories. Adding rolling embeddings to directed graph features significantly degrades primary accuracy (difference 0.018; 95\% CI $[0.004, 0.030]$; Holm-adjusted $p = 0.027$), while the rolling-versus-static comparison is inconclusive ($p = 0.616$). Moreover, the zero baseline achieves the lowest pooled log-MAE (1.391), and rolling-model $R^2$ is 0.0034. Held-out TreeSHAP audits across 2,500 observations attribute 67.24\% of raw $\log(1+y)$ output magnitude to own histories and booked maturities, 23.51\% to directed graph features, and 9.24\% to aligned rolling embeddings. These results establish a transparent, leakage-audited negative benchmark for contractual invoice forecasting under resource-bounded temporal graph representations.

\end{abstract}

\section{Introduction}
\label{sec:intro}

Financial forecasting changes substantive meaning when the empirical observations are contractual invoices rather than realized payment streams. An invoice face value and contractual due date record a bilateral commercial claim; they do not establish whether that claim remained unpaid, was settled early, was renegotiated, or ultimately entered default. This distinction anchors the present study because the analyzed ledger provides directed debtor--creditor links, exact integer euro-cent amounts, and literal calendar issue and due dates, without settlement timestamps or default flags \citep{esteva2026dataset}. Consequently, our scientific inquiry asks whether past relational structure and temporal graph embeddings improve out-of-sample forecasts of subsequently recorded invoice issuance and contractual maturity exposure, rather than estimating borrower probability of default or realized cash flows.

Trade credit links liquidity conditions across counterparties in interfirm production and supply chains. Seminal corporate-finance research links reliance on supplier finance to constrained institutional credit and documents how unconstrained firms intermediate liquidity along credit chains \citep{petersen1997trade,kiyotaki1997credit,boissay2013payment}. Empirical evidence further shows that supplier bargaining power shapes both the volume and contractual terms of trade credit when payment histories and product-market covariates are observed \citep{fabbri2016bargaining}. At the macroeconomic level, asymmetric production and trade-credit networks transmit firm-level shocks into aggregate fluctuations \citep{acemoglu2012network,carvalho2014micro}. While our eight-field invoice ledger cannot identify structural financing constraints or bargaining mechanisms, these economic asymmetries explain why debtor-side (outgoing) and creditor-side (incoming) histories should never be collapsed into an undirected firm aggregate.

Earlier relational studies demonstrate that network features can improve financially labeled default predictions when external credit outcomes are available. \citet{berloco2021predicting} combined firm covariates with a bank-observed transaction and delayed-payment network to forecast integrated corporate default over a three-month horizon, reporting improved recall relative to non-network baselines. However, observed bank-payment networks contain settlement behavior absent from pure invoice ledgers, so classification gains from default studies cannot be transferred without empirical verification to regression forecasts of new contractual issuance.

\begin{figure}[tbp]
  \centering
  \includegraphics[width=\textwidth]{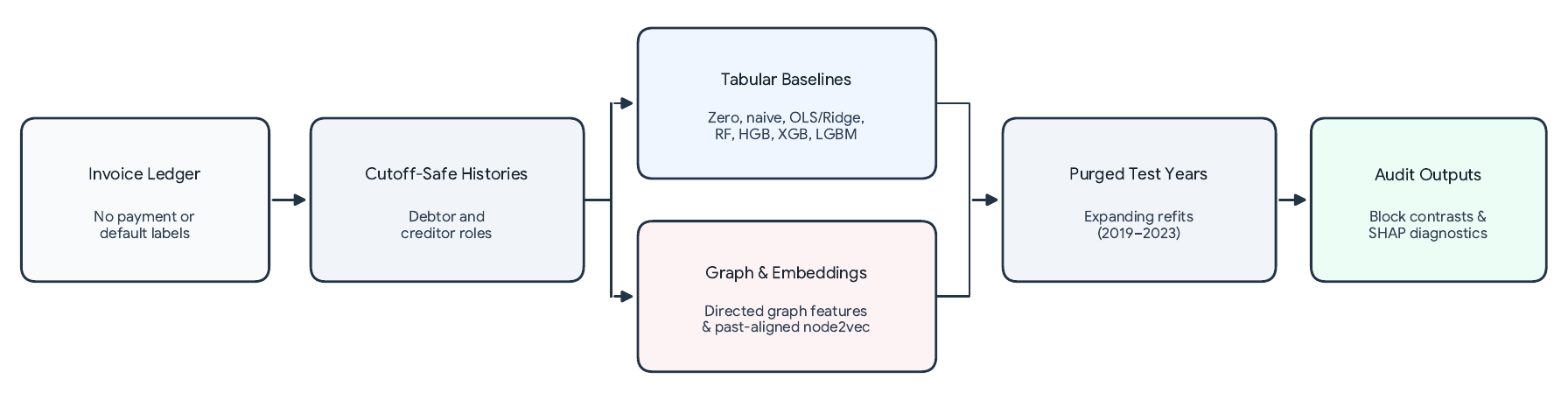}
  \caption{End-to-end empirical benchmarking and validation framework. Solid arrows denote computational processing stages rather than causal economic mechanisms.}
  \label{fig:research-framework}
\end{figure}

In prior companion research \citep{madugula2026beyond}, we audited the same 2012--2023 SME receivables ledger, analyzed temporal network topology and maturity concentration, and introduced the Maturity-Aware Network Exposure (MANE) indicator for systemic liquidity exposure. That study focused on structural characterization, maturity concentration, and contemporaneous network associations rather than leakage-purged multi-learner predictive benchmarking. The present study extends \citet{madugula2026beyond} into an explicit predictive setting: it separates cutoff-known booked maturities from genuinely future 30-, 60-, and 90-day issuance targets, reuses the audited cent-preserving graph engine and MANE construction strictly as cutoff-safe candidate predictors, benchmarks fourteen model and feature-block configurations under horizon-purged expanding validation, quantifies empirical prediction-interval coverage, and audits held-out TreeSHAP attribution stability. Crucially, network complexity is tested rather than presupposed; Figure~\ref{fig:research-framework} outlines the end-to-end benchmarking protocol, explicitly accommodating null or adverse incremental graph effects.

\section{Research Background and Literature Review}
\label{sec:literature}

\subsection{Relational Finance and Closest Forecasting Studies}
\label{sec:lit-relational}

Financial graph learning maps commercial interactions onto edges while attaching prediction targets to nodes, edges, or paths \citep{rajput2022temporal}. Despite shared graph formalism, task labels differ fundamentally across financial applications. Temporal guarantee-loan studies forecast binary delinquency cascades \citep{cheng2020delinquent}, causal-anonymous-walk frameworks benchmark transaction-based credit scoring \citep{makarov2022temporal}, and time-variant contrastive architectures model neighbor propagation for corporate credit risk and financial distress \citep{wang2025modeling}. Our forecasting horizons instead target continuous future invoice counts, issued face values, and newly issued within-horizon contractual maturities.

Among receivables-specific forecasting studies, \citet{kureljusic2023applicability} analyzed over one million invoices from a single DAX-40 enterprise during 2017--2019 to forecast customer payment dates on a daily rolling basis, highlighting the strong competitiveness of naive baselines against machine learning and deep learning models. Unlike that single-enterprise cash-collection setting, our active-firm panel spans 75,343 pseudonymous participants across a multi-firm network and lacks observed payment dates.

Explainable credit-scoring literature further clarifies target boundaries. \citet{delange2022explainable} and \citet{melsom2022explainable} combined LightGBM with Shapley attributions to predict default on unsecured bank loans, while \citet{bussmann2020explainable} grouped SME credit borrowers via similarity in Shapley explanations. These applications establish practical precedent for post-hoc tree explanations in finance, yet contractual invoice forecasts are neither credit scores nor fairness-audited underwriting tools, particularly because no protected borrower attributes exist in the ledger.

\subsection{Information Retained and Discarded by Graph Representations}
\label{sec:lit-graph}

Structural indicators capture distinct network properties. Classical graph theory separates degree, closeness, and betweenness \citep{freeman1978centrality}, while exchange-network theory demonstrates that centrality does not automatically equal bargaining power \citep{bonacich1987power}. Weighted-network generalizations balance tie counts against transaction volumes \citep{opsahl2010node}, whereas PageRank originated in hypertext ranking rather than financial loss modeling \citep{brin1998anatomy}. Directed clustering distinguishes cycle and transitive triangle orientations \citep{fagiolo2007clustering}, and weighted reciprocity requires preserving monetary magnitudes rather than binary reverse-link indicators \citep{squartini2013reciprocity}. Dynamic-network surveys emphasize that aggregating temporal edges alters reachability and temporal motifs \citep{holme2012temporal}, while walk-based dynamic centrality explicitly discounts edge age and separates temporal ordering from static topology \citep{arrigo2022dynamic,colman2016separating}.

Representation learning maps graph neighborhoods into continuous vector spaces. Early transductive random-walk methods such as DeepWalk \citep{perozzi2014deepwalk} and biased-walk \texttt{node2vec} \citep{grover2016node2vec} preserve local and structural proximity, while neural message-passing architectures including Graph Convolutional Networks \citep{kipf2017semi}, GraphSAGE \citep{hamilton2017inductive}, and Graph Attention Networks \citep{velickovic2018graph} aggregate node features inductively. Dynamic graph surveys \citep{kazemi2020representation} categorize extensions ranging from aligned snapshot embeddings \citep{singer2019node} and self-attentive DySAT \citep{sankar2018dynamic} to continuous-time memory architectures such as Temporal Graph Networks \citep{rossi2020temporal}. Related temporal paradigms include multivariate time-series graph learning \citep{wu2020connecting} and logical-rule temporal knowledge-graph forecasting \citep{liu2022tlogic}.

Crucially, empirical benchmarks warn against assuming graph superiority. The Temporal Graph Benchmark \citep{huang2023temporal} reports substantial cross-dataset variance and demonstrates that simple node-property baselines frequently rival complex temporal graph architectures on dynamic node-property tasks. Our benchmark implements compact, cutoff-safe rolling \texttt{node2vec} embeddings aligned across consecutive origins via orthogonal Procrustes rotation \citep{schonemann1966generalized} and fed into a fixed gradient-boosted learner. This design intentionally isolates whether lightweight aligned embeddings add value over explicit directed graph features; it is not an event-memory neural TGN, and our undirected projection deliberately discards edge direction in the embedding block---a trade-off probed directly through directional feature ablations.

\subsection{Forecasting, Validation, and Explanation as Distinct Problems}
\label{sec:lit-eval}

Comparative tabular machine-learning studies establish that gradient-boosted decision trees remain highly competitive with deep neural networks on heterogeneous tabular data, with no universally dominant model family \citep{shwartzziv2022tabular,grinsztajn2022why,gorishniy2021revisiting}. On heavy-tailed, zero-inflated responses, error-metric selection is consequential: consistent scoring functions link loss choice directly to target functionals \citep{gneiting2011making}, while per-row percentage errors become degenerate at zero \citep{hyndman2006another}. Furthermore, inverting log-transformed mean predictions without retransformation bias correction \citep{duan1983smearing} does not yield an unbiased conditional mean on the original monetary scale, explaining why log-scale and raw-EUR rankings can diverge.

Under nonstationarity, standard random cross-validation leaks temporal structure and inflates apparent accuracy. Out-of-sample rolling-origin designs \citep{tashman2000outofsample}, blocked evaluation \citep{bergmeir2012use}, and repeated prequential testing \citep{cerqueira2020evaluating} provide more reliable performance estimates, directly addressing leakage pitfalls cataloged in scientific machine learning \citep{kapoor2023leakage}. Similarly, uncertainty quantification via conformal prediction \citep{vovk2005algorithmic,lei2018distribution,romano2019conformalized} guarantees finite-sample marginal coverage under exchangeability, whereas nonexchangeable time series motivate weighted, ensemble, adaptive, or relational conformal extensions \citep{barber2023conformal,xu2023conformal,gibbs2021adaptive,zhang2025uncertainty,cini2025relational}. Because our benchmark applies fixed pre-test split-residual quantiles without online adaptation, we treat interval coverage strictly as an empirical diagnostic.

Finally, post-hoc explainability must not be conflated with causal identification or accuracy attribution. Rooted in cooperative game theory \citep{shapley1953value}, SHAP and TreeSHAP provide consistent local additive attributions for tree ensembles \citep{lundberg2017unified,lundberg2020local} within broader responsible-AI taxonomies \citep{barredo2020explainable}. Yet feature correlation alters interventional versus conditional attributions \citep{aas2021explaining}, axiomatic Shapley properties do not automatically satisfy human causal intuition \citep{kumar2020problems}, post-hoc explanations cannot substitute for inherently interpretable high-stakes systems \citep{rudin2019stop}, and recent credit-scoring decomposition research proves that features driving individual predictions need not coincide with features driving aggregate predictive accuracy \citep{hue2026measuring}.

\section{Research Contribution and Prespecified Hypotheses}
\label{sec:hypotheses}

This study contributes an audited, leakage-safe forecasting benchmark for contractual trade-credit ledgers that: (i) separates cutoff-known booked maturities from genuinely future 30-, 60-, and 90-day issuance targets; (ii) evaluates incremental relational and temporal-embedding value under an identical XGBoost learner against strong zero, naive, linear, and tree baselines; and (iii) jointly reports multi-scale error metrics, calendar stability, empirical 90\% prediction-interval coverage, and held-out TreeSHAP attribution stability. Rather than claiming universal originality or guaranteed graph gains, we frame four explicit research hypotheses:

\begin{itemize}
  \item \emph{Hypothesis H1 (Full rolling graph vs.\ tabular baseline):} Appending directed graph features and past-only aligned rolling \texttt{node2vec} embeddings improves primary forecasting accuracy relative to non-graph tabular XGBoost.
  \item \emph{Hypothesis H2 (Incremental rolling embeddings vs.\ directed graph features):} Adding aligned rolling \texttt{node2vec} embeddings improves primary forecasting accuracy beyond tabular histories plus explicit directed graph features.
  \item \emph{Hypothesis H3 (Rolling vs.\ prior-year static embeddings):} Rolling monthly aligned embeddings outperform previous-complete-year static embeddings when tabular histories, directed graph features, and learner hyperparameters are held constant.
  \item \emph{Hypothesis H4 (Explanatory family differentiation):} Own-firm invoice histories/booked maturities, directed graph features, and aligned rolling embeddings exhibit systematically different global TreeSHAP attribution shares and cross-year rank stability in the primary forecasting model.
\end{itemize}

The prespecified primary endpoint is debtor-side newly issued face value whose contractual due dates fall within the next 30 calendar days. Inferential testing for H1--H3 evaluates equal-date mean absolute error on $\log(1 + \text{EUR})$, where negative temporal-minus-comparator differences favor the rolling temporal representation. Hypotheses H1--H3 constitute a three-test family controlled via Holm's sequentially rejective procedure \citep{holm1979simple}. H4 is evaluated descriptively via global feature-family shares and cross-year rank correlations rather than as a causal significance test. Count, face-value, creditor-side, 60/90-day, and network-total tasks serve as supplementary descriptive benchmarks.

\section{Dataset Provenance, Integrity Audit, and Variable Definitions}
\label{sec:dataset}

\subsection{Source Identity and Analytical Population}
\label{sec:source-id}

The analyzed master ledger (\code{pseudonymized_atomic_invoice_records_2012_2024-02-08.csv}) is sourced from the Mendeley Data repository associated with our prior trade-credit network investigation \citep{esteva2026dataset,madugula2026beyond} (CC BY 4.0; companion curation repository at \url{https://github.com/peplluis7/onemillioninvoices}). It comprises eight fields per invoice: unique invoice pseudonym, debtor pseudonym, creditor pseudonym, integer face value in euro cents, literal ISO issue date, literal ISO contractual due date, issue cohort year, and analysis role. Cryptographic verification confirms the exact master-file digest:
\[
  \text{SHA-256} = \texttt{\seqsplit{e139b8dc75b7334d243874b6ebdaf1798c9969038b193f7eb4cd0e873bcc1a88}}.
\]
Annual split files distributed alongside the master serve strictly for reconciliation and are never concatenated as additional observations.

Full-ledger programmatic auditing retains exactly 749,952 analytical invoices issued during 2012--2023 totaling 9,970,501,556,950 euro cents (EUR~99.705 billion) across 75,343 distinct analytical firms (union of debtors and creditors). The audit isolates and excludes 1,240 terminal 2024 bridge records totaling 63,011,692,269 cents and nine bridge-only firms (ensuring the full-stream union of 75,352 firms is never conflated with the analytical population). Table~\ref{tab:data-provenance} summarizes provenance, scope, and unobserved economic outcomes.

\begin{table}[tbp]
  \centering
  \caption{Data provenance, cohort partitioning, and verification boundaries for the audited master invoice ledger.}
  \label{tab:data-provenance}
  \begin{adjustbox}{max width=\textwidth}
  \begin{tabular}{@{}lll@{}}
    \toprule
    \textbf{Audit Component} & \textbf{Identity and Coverage} & \textbf{Treatment and Verification} \\
    \midrule
    Analytical master file & \code{pseudonymized_atomic_invoice_records_2012_2024-02-08.csv} & SHA-256 audited; loaded once; zero row deduplication override \\
    Repository reference & Mendeley Data Version 1 \citep{esteva2026dataset} & DOI \code{10.17632/28rbmvwsm9.1}; aligns with \citet{madugula2026beyond} \\
    Main analytical cohort & Recorded 2012--2023 issue cohorts & 749,952 rows; 75,343 union firms; EUR~99.705 billion exact cents \\
    Terminal follow-up & Recorded 2024 \code{terminal_bridge_only} role & 1,240 rows (9 bridge-only firms) excluded from features and targets \\
    Annual CSV splits & 2012--2024 annual release files & Reconciled against master totals only; never appended \\
    Unobserved outcomes & Payments, default, recoveries, cash receipts, firm size & Never inferred, imputed, or simulated \\
    \bottomrule
  \end{tabular}
  \end{adjustbox}
\end{table}

Repository documentation notes heterogeneous multi-source curation and a separately curated 2021 corpus without a public entity crosswalk \citep{madugula2026beyond,esteva2026dataset}. Pseudonyms are therefore processed literally; apparent firm exit or re-entry across calendar years may combine true economic dynamics with curation shifts. Because firm headcount and revenue are unobserved, ``SME'' reflects the institutional receivables setting rather than a row-level verified size classification.

\subsection{Validation Checks, Literal Dates, and Monetary Precision}
\label{sec:validation-checks}

Programmatic schema validation verified zero missing required fields, zero duplicate invoice identifiers, zero nonpositive amounts, zero self-invoiced debtor--creditor pairs, zero negative contractual terms ($u_i < s_i$), and zero issue-year cohort mismatches across all 749,952 analytical records. Records sharing economic attributes under distinct invoice pseudonyms are preserved verbatim. Consistent with \citet{madugula2026beyond}, contractual face values exhibit extreme right-skewness: mean EUR~132,948.53, median EUR~1,272.35, 99th percentile EUR~1,239,060.41, maximum EUR~1,921,029,610.00, Gini coefficient 0.9742, with the top 1\% of invoices representing 81.41\% of total face value. Contractual terms have median 10 days and 90th percentile 60 days, while 13 records exceed 3,650 days (maximum recorded due date 5656-03-08). Rather than applying ad-hoc date repairs, literal ISO dates are converted to integer calendar days to avoid 64-bit nanosecond timestamp overflow.

\begin{figure}[tbp]
  \centering
  \includegraphics[width=\textwidth]{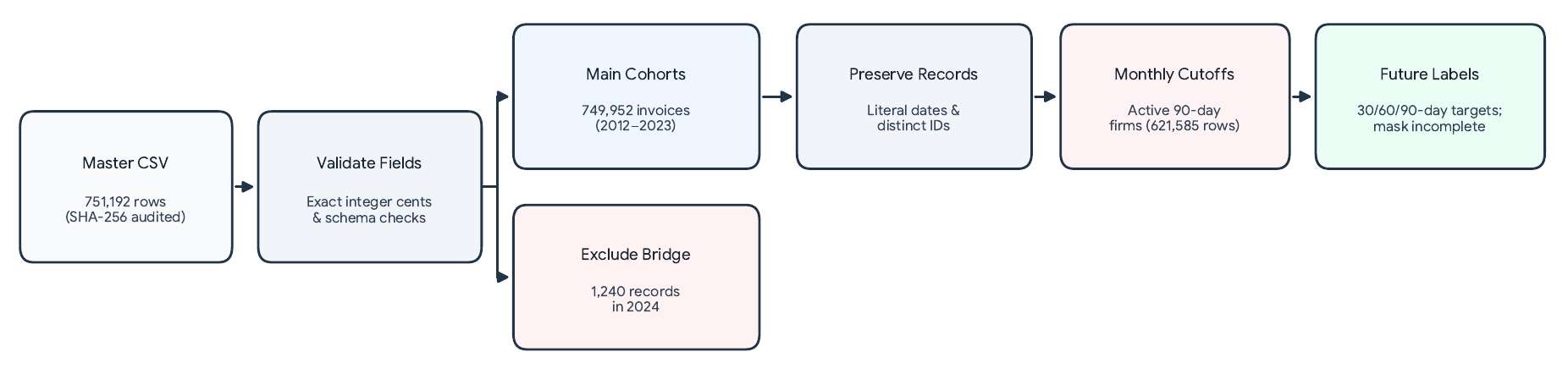}
  \caption{Executed master-data validation, cohort filtering, and temporal cutoff pipeline across the 751,192 raw rows and 749,952 analytical invoices.}
  \label{fig:data-pipeline}
\end{figure}

Integer cents remain authoritative for source storage and exact ledger totals. Because the retained nonnegative cent mass ($9.97 \times 10^{12}$) lies strictly below $2^{53} \approx 9.01 \times 10^{15}$, unweighted nonnegative bincount aggregations in double precision are exact prior to integer conversion on this ledger (a dataset-specific property rather than a general guarantee). Subsequent division by 100, exponential kernels, graph normalization, tree regression, and TreeSHAP computations proceed in standard IEEE-754 floating-point arithmetic. Figure~\ref{fig:data-pipeline} documents the executed ingestion, filtering, and cutoff-labeling pipeline.

\section{Prediction-Target Construction}
\label{sec:targets}

Let each invoice $i$ carry debtor $d_i$, creditor $c_i$, integer euro-cent face value $a_i > 0$, integer issue day $s_i$, and integer due day $u_i \ge s_i$. For a monthly forecast origin (cutoff) $t$ and forecast horizon $h \in \{30, 60, 90\}$ calendar days, define the set of newly issued invoices strictly after the cutoff and within the horizon, together with its subset maturing within the same horizon:
\begin{align}
  N(t, h) &= \{\, i : t < s_i \le t + h \,\}, \label{eq:new-set} \\
  M(t, h) &= \{\, i \in N(t, h) : u_i \le t + h \,\}. \label{eq:mature-set}
\end{align}
Because schema validation enforces $u_i \ge s_i$, every invoice $i \in M(t, h)$ satisfies $t < s_i \le u_i \le t + h$. For directional role $r \in \{\mathrm{out}, \mathrm{in}\}$, let $I_i^{\mathrm{out}}(v) = \mathbb{I}(d_i = v)$ and $I_i^{\mathrm{in}}(v) = \mathbb{I}(c_i = v)$. The three firm-level prediction target families are defined as:
\begin{align}
  Y_{v,t,h}^{\mathrm{count}, r} &= \sum_{i \in N(t, h)} I_i^r(v), \qquad
  Y_{v,t,h}^{\mathrm{amount}, r} = \sum_{i \in N(t, h)} \frac{a_i}{100}\, I_i^r(v), \label{eq:firm-count-amt} \\
  Y_{v,t,h}^{\mathrm{mature}, r} &= \sum_{i \in M(t, h)} \frac{a_i}{100}\, I_i^r(v). \label{eq:firm-mature}
\end{align}
Crucially, already-issued claims that mature during $(t, t+h]$ are known at cutoff $t$ and therefore belong exclusively to the predictor set rather than the uncertain target:
\begin{equation}
  B_{v,t,h}^{r} = \sum_{\substack{i:\, s_i \le t \\ t < u_i \le t + h}} \frac{a_i}{100}\, I_i^r(v).
  \label{eq:booked-maturity}
\end{equation}
Separating $B_{v,t,h}^{r}$ from $Y_{v,t,h}^{\mathrm{mature}, r}$ prevents target leakage from pre-existing contractual schedules while allowing every tabular baseline full access to booked maturities.

At the aggregate network level, future issuance counts and face values count each newly issued invoice exactly once across the entire ledger, including invoices from future-entering firms. Network maturity-weighted exposure applies a 30-day exponential proximity kernel:
\begin{equation}
  Y_{t,h}^{\mathrm{weighted}} = \sum_{i \in M(t, h)} \frac{a_i}{100} \exp\!\left(-\frac{u_i - t}{30}\right).
  \label{eq:network-weighted}
\end{equation}
This EUR-scaled proximity kernel is a defined temporal-concentration metric rather than a credit-loss discount factor. Summing firm-level debtor and creditor exposures would double-count bilateral claims and is never used as a network denominator.

Forecast origins run monthly from 31 January 2013 through 30 November 2023 (131 origins). A target label at $(t, h)$ is defined only if $t + h \le \text{2023-12-31}$; incomplete terminal horizons are masked as missing (\code{NaN}) rather than coded as zero. The firm-level forecasting population at origin $t$ comprises all firms with at least one recorded invoice in either role during the trailing 90 calendar days $(t-90, t]$. Because this active-set rule depends solely on past issues, it introduces zero future-selection leakage and yields exactly 621,585 firm-origin observations across 75,266 distinct active firms. Table~\ref{tab:target-families} summarizes all observed and derived target families.

\begin{table}[tbp]
  \centering
  \caption{Observed master-ledger fields and derived firm-level and network-level prediction target families.}
  \label{tab:target-families}
  \begin{adjustbox}{max width=\textwidth}
  \begin{tabular}{@{}lllll@{}}
    \toprule
    \textbf{Variable Family} & \textbf{Source Fields} & \textbf{Analytical Grain} & \textbf{Unit} & \textbf{Missingness Rule} \\
    \midrule
    Observed invoice fields & All 8 master columns & Literal recorded invoice & Cents, dates, IDs & Zero missingness allowed \\
    Future issuance count & Issue date, role IDs & Firm--role--horizon ($t < s_i \le t+h$) & Count & Empty complete = 0; incomplete = NaN \\
    Future issued face value & Above + amount & Firm--role--horizon ($t < s_i \le t+h$) & EUR & Empty complete = 0; incomplete = NaN \\
    New maturity exposure & Above + due date & Firm--role--horizon ($i \in M(t,h)$) & EUR & Empty complete = 0; incomplete = NaN \\
    Booked maturities (predictor) & Amount, dates, role IDs & Firm--role--horizon ($s_i \le t < u_i \le t+h$) & EUR & Empty booked schedule = 0 \\
    Network weighted exposure & Amount, issue \& due dates & System origin--horizon (counted once) & EUR-scaled & Incomplete horizon = NaN \\
    External / simulated covariates & None & None fitted & Not applicable & Not applicable \\
    \bottomrule
  \end{tabular}
  \end{adjustbox}
\end{table}

\section{Feature Engineering}
\label{sec:features}

\begin{figure}[tbp]
  \centering
  \includegraphics[width=\textwidth]{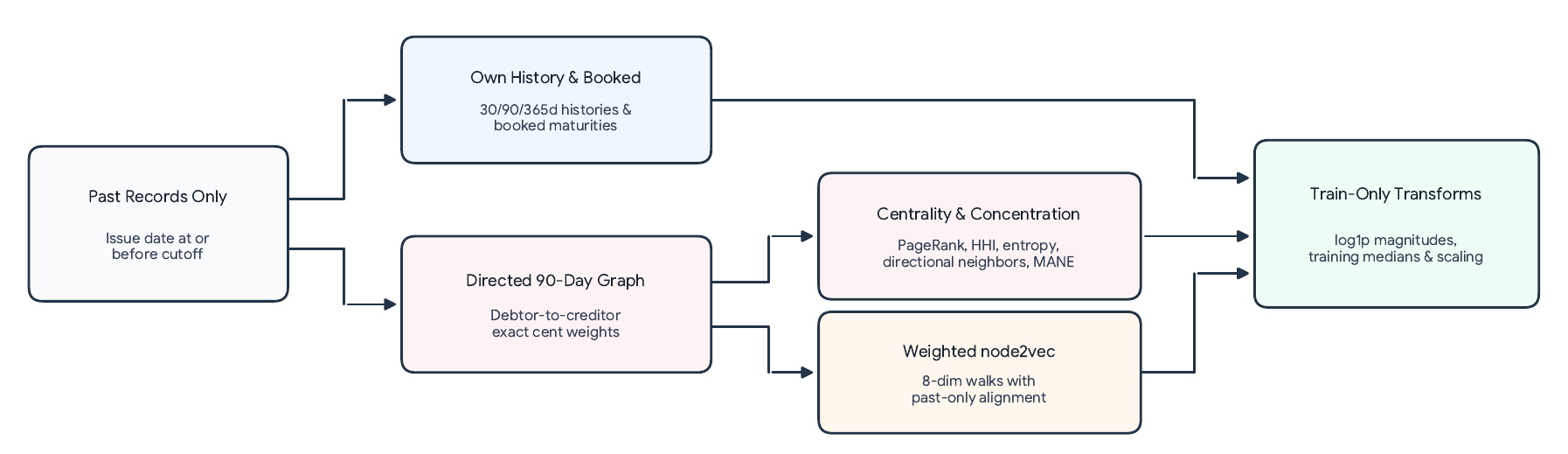}
  \caption{Cutoff-safe feature architecture combining own-firm histories and booked maturities, directed 90-day graph features, past-aligned \texttt{node2vec} embeddings, and training-only preprocessing.}
  \label{fig:feature-architecture}
\end{figure}

All predictors at cutoff $t$ restrict invoice inputs strictly to $s_i \le t$. Features are organized into three nested blocks so incremental representational value can be isolated cleanly under a fixed learner (Figure~\ref{fig:feature-architecture}).

\subsection{Tabular Own-Firm Block}
\label{sec:feat-tabular}

For each active firm $v$ and role $r \in \{\mathrm{out}, \mathrm{in}\}$, own histories aggregate 30-, 90-, and 365-day trailing issued counts, issued face values, and recently issued invoices whose due dates also fell within the corresponding trailing window (\code{hist_mature}). The block further includes cutoff-known booked maturities $B_{v,t,h}^r$ for $h \in \{30, 60, 90\}$, firm age in days since first observed invoice (\code{age_days}), cyclical calendar-month encodings ($\sin$ and $\cos$), mean remaining maturity across open claims, and direct proximity seeds:
\begin{equation}
  q_{v,t}^r = \sum_{\substack{i:\, t-90 < s_i \le t \\ u_i \ge t}} \frac{a_i}{100}\, I_i^r(v) \exp\!\left(-\frac{u_i - t}{30}\right).
  \label{eq:proximity-seed}
\end{equation}
Thus, non-graph tabular baselines already receive rich directional histories and booked maturity schedules; relational blocks must add predictive signal beyond these controls.

\subsection{Directed 90-Day Graph Block}
\label{sec:feat-graph}

At each monthly origin $t$, we construct the weighted directed graph $G_t = (V_t, E_t, W_t)$ from all invoices issued during $(t-90, t]$, with debtor-to-creditor edge weights equal to exact aggregated euro cents. From $G_t$ we compute directional degrees, weighted PageRank, counterparty Herfindahl--Hirschman concentration (HHI) and Shannon entropy, 7-day-bin future maturity HHI, joint counterparty--maturity HHI, and role balance $2\min(s_v^{\mathrm{out}}, s_v^{\mathrm{in}})/(s_v^{\mathrm{out}} + s_v^{\mathrm{in}})$. Empty role-specific maturity schedules yield missing HHI or mean remaining days rather than artificial zeros. First-order neighbor predictors multiply the row-normalized adjacency matrix $C_t^{\mathrm{out}}$ and its transposed creditor-side analogue $C_t^{\mathrm{in}}$ by historical activity and proximity seeds. Reusing the audited recursion from \citet{madugula2026beyond} purely as a candidate predictor, directional temporal exposure evolves via
\begin{equation}
  z_t^r = (1 - \alpha_t)\, q_t^r + \alpha_t\, C_t^r\, z_{t-\Delta t}^r, \qquad
  \alpha_t = 0.5 \exp\!\left(-\frac{\Delta t}{90}\right),
  \label{eq:mane-recursion}
\end{equation}
with combined indicator $\mathrm{MANE}_t = \tfrac{1}{2}(z_t^{\mathrm{out}} + z_t^{\mathrm{in}})$.

\subsection{Preprocessing and System-Level Blocks}
\label{sec:feat-preproc}

Nonnegative magnitudes (histories, booked schedules, neighbor aggregates, age, remaining maturity, seeds, MANE, and degrees) undergo $\log(1+x)$ transformation, while PageRank, HHI, entropy, role balance, cyclical terms, and signed embeddings remain on their natural scales. Undefined values are imputed using training-fold medians only (all-missing training columns default to zero). For aggregate network-level forecasting, the tabular block includes lagged system counts, face values, kernel exposures, cyclical calendar features, and global seed; the system graph block adds active node/edge counts, density, weighted reciprocity, binary edge Jaccard persistence, directional mean maturity HHI, dual-role share, and mean outgoing degree.

\section{Temporal Graph Representation Methodology}
\label{sec:temporal-graph}

To test whether low-dimensional latent graph coordinates improve forecasting beyond explicit directed topological features, we generate monthly rolling \texttt{node2vec} embeddings \citep{grover2016node2vec} from the symmetrized undirected weighted projection $U_t = W_t + W_t^\top$, where reciprocal bilateral volumes sum and edge direction is discarded exclusively within the embedding block. Weighted biased random walks use return parameter $p = 1$, in-out parameter $q = 2$, 2 walks per active node, walk length 12, embedding dimension $d = 8$, skip-gram context window 4, 5 negative samples, and 5 training epochs. Exact biased transition probabilities are sampled via cached cumulative-weight proposals and rejection sampling to prevent quadratic hub scans. Skip-gram training executes deterministically with a single worker, fixed seed 1729, and stable token hashing.

Because skip-gram coordinate frames are rotationally invariant across snapshots, raw embeddings at origin $t$ ($E_t \in \mathbb{R}^{|V_t| \times 8}$) are $\ell_2$-normalized per node and orthogonally aligned to the immediately preceding monthly frame $E_{t-1}$ whenever at least eight shared active firms exist ($S_t = V_t \cap V_{t-1}$, $|S_t| \ge 8$). Following \citet{schonemann1966generalized} and \citet{singer2019node}, the orthogonal rotation matrix $R_t^* \in O(8)$ minimizes Frobenius discrepancy strictly over past-known shared nodes:
\begin{equation}
  R_t^* = \arg\min_{R^\top R = I_8} \left\| E_{t, S_t} R - E_{t-1, S_t} \right\|_F^2.
  \label{eq:procrustes}
\end{equation}
Alignment uses only current and prior snapshots and never accesses future graphs or labels. As a static-representation comparator for H3, we construct embeddings from the previous complete calendar year (causally aligned across prior years); active firms absent from the previous calendar year receive zero coordinates accompanied by an explicit availability flag, whereas rolling embedding availability equals 1 for all active firms.

\begin{figure}[tbp]
  \centering
  \includegraphics[width=\textwidth]{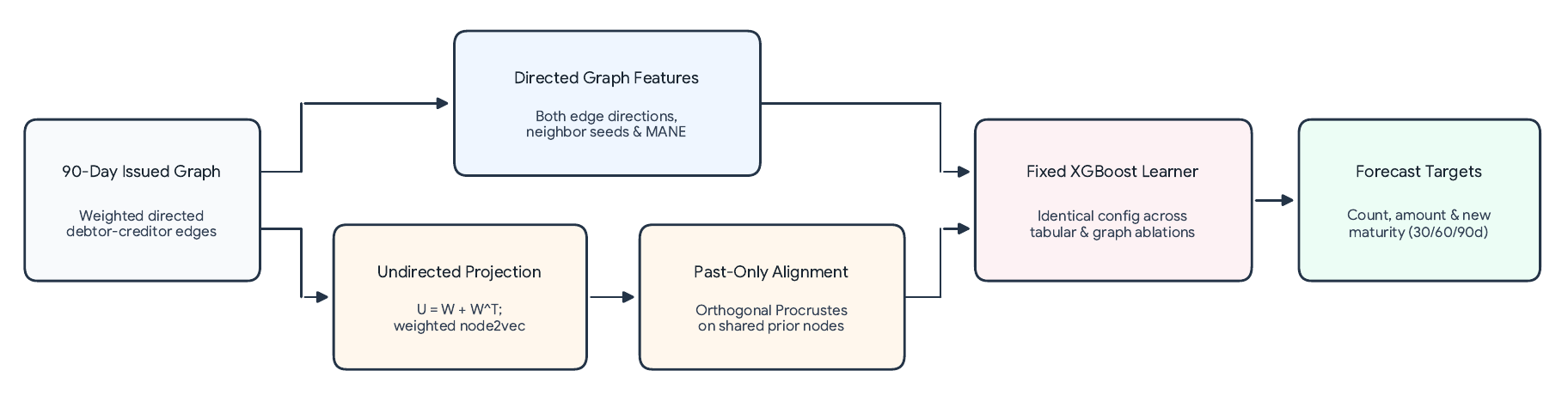}
  \caption{Implemented temporal graph machine-learning architecture combining directed graph features, undirected weighted \texttt{node2vec}, past-only orthogonal Procrustes alignment, and a fixed XGBoost learner.}
  \label{fig:graph-ml-architecture}
\end{figure}

As illustrated in Figure~\ref{fig:graph-ml-architecture}, the graph-feature comparator pairs tabular histories with explicit directed graph features; the static and rolling comparators append their respective 8-dimensional embeddings to that exact block under identical XGBoost settings. Partial directional ablations (\code{no_debtor} and \code{no_creditor}) drop graph columns carrying outgoing or incoming directional suffixes while retaining own-firm histories, shared-role features, and undirected embeddings. Notably, this pipeline contains no neural event-memory TGN, learned end-to-end message passing, smart-contract layer, or decentralized training component.

\section{Machine Learning Models and Experimental Design}
\label{sec:methods}

\subsection{Learners and Fixed Hyperparameter Configurations}
\label{sec:learners}

We evaluate three analytical baselines alongside six supervised tabular learners and five relational/ablation XGBoost variants without accuracy-driven hyperparameter tuning (Table~\ref{tab:model-configs}). Baselines comprise constant zero, the trailing 30-day activity rate scaled by $h/30$, and the trailing 365-day historical rate scaled by $h/365$. Linear models comprise ordinary least squares (OLS) and ridge regression ($\alpha = 10$) \citep{hoerl1970ridge} on standardized inputs (Lasso \citep{tibshirani1996regression} is not fitted). Nonlinear tree ensembles comprise Random Forests \citep{breiman2001random}, Histogram Gradient Boosting \citep{friedman2001greedy}, XGBoost \citep{chen2016xgboost}, and LightGBM \citep{ke2017lightgbm}.

\begin{table}[tbp]
  \centering
  \caption{Executed learner and feature-block configurations fixed prior to held-out evaluation.}
  \label{tab:model-configs}
  \begin{adjustbox}{max width=\textwidth}
  \begin{tabular}{@{}ll@{}}
    \toprule
    \textbf{Model Family} & \textbf{Fixed Executed Configuration} \\
    \midrule
    Zero / naive / historical rate & Constant 0; trailing 30-day rate scaled by $h/30$; trailing 365-day rate scaled by $h/365$ \\
    OLS / Ridge & Training-only \code{StandardScaler}; ordinary least squares / Ridge ($\alpha = 10$) \\
    Random Forest & 60 trees; max depth 10; min leaf 10; feature fraction 0.8; subsample 0.7 \\
    Histogram Gradient Boosting & 120 iterations; 15 leaves; max depth 5; learning rate 0.05; $L_2 = 1$; no early stopping \\
    XGBoost (tabular \& graph variants) & 160 trees; max depth 3; learning rate 0.05; min child weight 10; $L_2 = 1$; \code{hist} solver \\
    LightGBM & 160 trees; max depth 4; 15 leaves; learning rate 0.05; min child samples 20; $L_2 = 1$ \\
    Graph / static / rolling / ablations & Identical XGBoost hyperparameters; only declared feature block varies \\
    Determinism \& resource cap & Seed 1729; up to 4 fit threads; 1 walk worker; max 5,000 sampled training rows/fold \\
    \bottomrule
  \end{tabular}
  \end{adjustbox}
\end{table}

All supervised learners optimize squared error on $z = \log(1 + Y)$, yielding raw predictions $\hat{z}$ that are clipped at zero and inverted via $\hat{Y} = \exp(\max(0, \hat{z})) - 1$. Because $\exp(\mathbb{E}[\log(1+Y)\mid X]) - 1$ is neither an unbiased conditional EUR mean without smearing retransformation \citep{duan1983smearing} nor optimized for raw-EUR loss, both log-scale and raw-EUR metrics are reported transparently.

\subsection{Purged Expanding-Window Validation}
\label{sec:validation}

\begin{figure}[tbp]
  \centering
  \includegraphics[width=\textwidth]{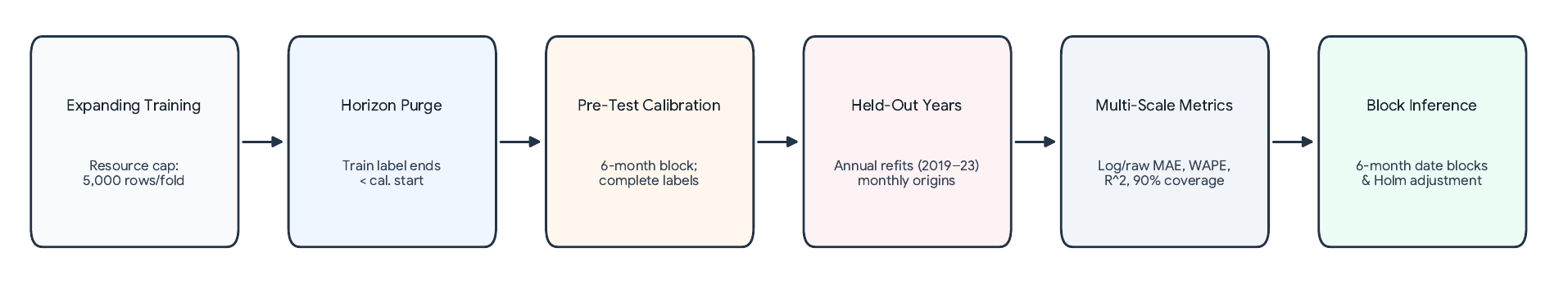}
  \caption{Purged expanding-window training, six-month pre-test residual calibration, and annual held-out rolling-origin evaluation across 2019--2023.}
  \label{fig:training-validation}
\end{figure}

Evaluation spans five held-out calendar years (2019--2023), covering 59 complete monthly origins (January 2019 through November 2023 for 30-day targets) and 606,641 held-out firm-origin rows. For each test year $Y_{\mathrm{test}}$, a six-month pre-test calibration block begins on 1 July of $Y_{\mathrm{test}} - 1$. Strict horizon purges enforce that every training observation's target window ends strictly before calibration starts ($t_{\mathrm{train}} + h < t_{\mathrm{cal,start}}$) and every calibration observation's target window ends strictly before 1 January of $Y_{\mathrm{test}}$ ($t_{\mathrm{cal}} + h < t_{\mathrm{test,start}}$). Calibration observations are never folded back into model fitting. Models refit annually on expanding historical windows while cutoff-known graphs and predictors update monthly throughout each test year (Figure~\ref{fig:training-validation}).

An initial unconstrained full-panel fit exhausted available system memory during late-year graph training; those incomplete runs were discarded. In the completed benchmark (820 total model fits in 438.14~s after 229.40~s panel construction), each eligible learned training fold is capped at 5,000 firm-origin rows sampled uniformly without replacement using seed $1729 + Y_{\mathrm{test}}$, while all complete calibration and test observations are retained intact. All comparators share identical training subsets, calibration rows, and test rows. Automated unit tests (\code{run_selftest.py}) verify prefix feature invariance against truncated ledgers, exact booked-versus-new separation, future-node exclusion, horizon masking, and zero identifier leakage.

\subsection{Metrics, Prediction Intervals, and Block Inference}
\label{sec:metrics-inference}

Performance is assessed across complementary scales: mean absolute log error ($\text{log-MAE} = n^{-1}\sum |\log(1+Y_i) - \log(1+\hat{Y}_i)|$), raw-unit MAE and RMSE, weighted absolute percentage error ($\text{WAPE} = \sum |Y_i - \hat{Y}_i| / \sum Y_i$), and out-of-sample $R^2 = 1 - \mathrm{SSE}/\mathrm{SST}$. Ordinary per-row MAPE is omitted due to pervasive zero targets \citep{hyndman2006another}.

Nominal 90\% prediction intervals derive from the finite-sample-adjusted 90th percentile $q_{0.90}$ of absolute log-scale residuals on the pre-test calibration block, mapped back to $[\exp(\max(0, \hat{z} - q_{0.90})) - 1,\, \exp(\max(0, \hat{z}) + q_{0.90}) - 1]$. Consistent with conformal prediction theory \citep{vovk2005algorithmic,lei2018distribution,barber2023conformal}, nonexchangeability and temporal shift preclude distribution-free coverage guarantees; coverage is reported strictly as empirical containment.

Primary hypothesis tests (H1--H3) compute equal-date mean log-MAE differences across the 59 held-out monthly origins, avoiding domination by cohorts with unusually large active-firm counts (such as 2020). Uncertainty intervals and two-sided centered $p$-values use 999 circular block bootstrap draws with 6-month blocks \citep{kunsch1989jackknife}. Unlike the stationary bootstrap \citep{politis1994stationary}, block lengths are fixed; unlike asymptotic serial-correlation tests \citep{diebold1995comparing} or multiway panel clustering \citep{cameron2011robust}, inference stays strictly tied to the executed date-block resampling distribution, supplemented by 3-month, 12-month, and raw-EUR sensitivity analyses. Family-wise error across H1--H3 is controlled via Holm adjustment \citep{holm1979simple}.

\section{Explainability Audit Protocol}
\label{sec:xai-method}

\begin{figure}[tbp]
  \centering
  \includegraphics[width=\textwidth]{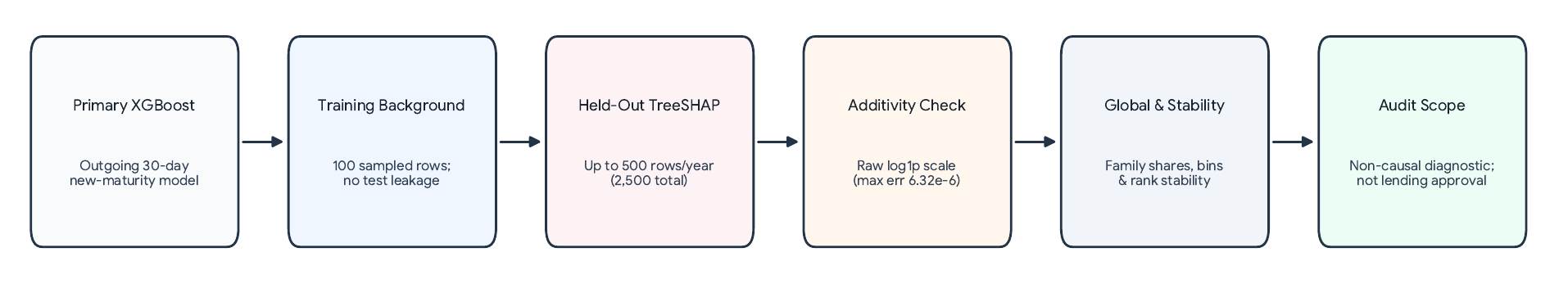}
  \caption{Executed held-out TreeSHAP audit workflow explaining raw $\log(1+\mathrm{EUR})$ model outputs without test-set background leakage.}
  \label{fig:explainability-workflow}
\end{figure}

Post-hoc explainability audits exclusively the primary rolling XGBoost model (outgoing 30-day new-maturity exposure) using interventional TreeSHAP \citep{lundberg2017unified,lundberg2020local}. For each held-out year (2019--2023), TreeSHAP evaluates up to 500 randomly sampled held-out test rows (2,500 total) against 100 randomly sampled rows drawn strictly from that fold's permitted training window (Figure~\ref{fig:explainability-workflow}). Attributions are computed on the unclipped raw $\log(1+\text{EUR})$ tree output prior to inverse transformation, ensuring local additivity can be verified numerically (maximum absolute additivity residual $6.32 \times 10^{-6}$ log units across all 2,500 rows).

Global importance aggregates mean absolute TreeSHAP values, feature-family shares sum mean absolute attributions across own-history/booked, directed-graph, and aligned-embedding predictors, and temporal stability measures pairwise Spearman rank correlations across held-out years. In line with recent critiques \citep{aas2021explaining,kumar2020problems,hue2026measuring}, TreeSHAP serves strictly as a model-behavior diagnostic rather than causal identification or lending-decision justification.

\section{Empirical Results and Model Comparisons}
\label{sec:results}

\subsection{Target Distributions and Temporal Network Shifts}
\label{sec:res-dist}

\begin{figure}[tbp]
  \centering
  \includegraphics[width=\textwidth]{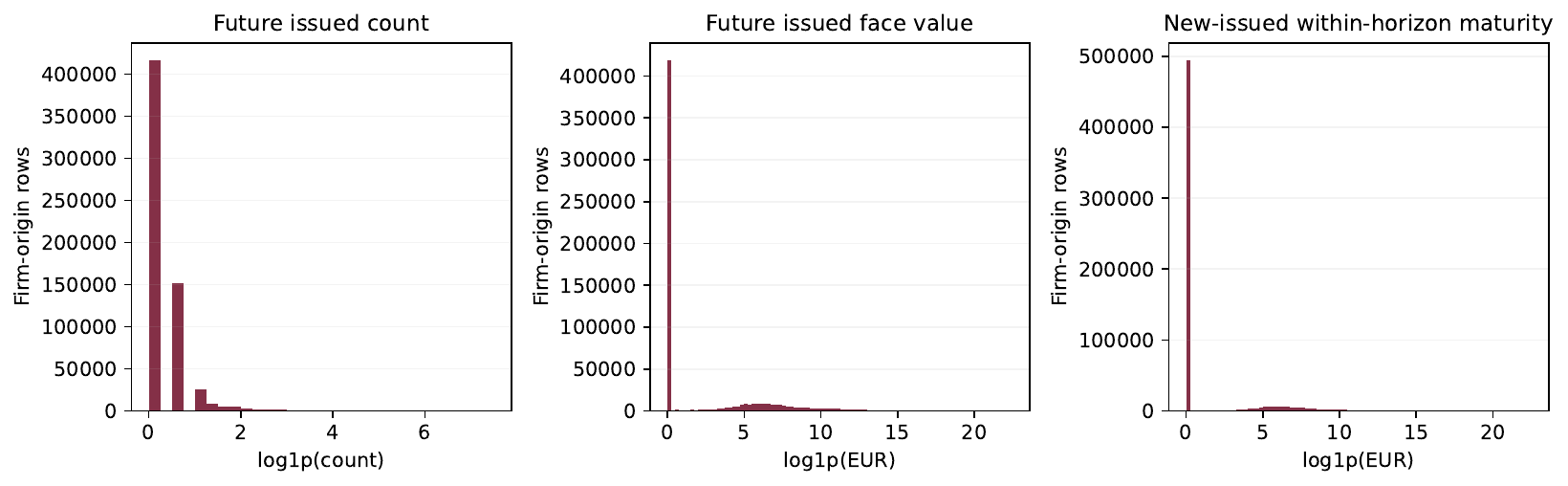}
  \caption{Distributions of 30-day debtor-side targets across complete firm-origin observations using 0.25-wide $\log(1+x)$ bins: future issued count, issued face value, and newly issued within-horizon maturity.}
  \label{fig:target-distributions}
\end{figure}

Across the 621,585 firm-origin panel observations (131 monthly origins, 75,266 distinct active firms), contractual activity is strongly zero-inflated and heavy-tailed (Figure~\ref{fig:target-distributions}). The primary outgoing 30-day new-maturity target equals zero in 79.28\% of complete firm-origin rows, with mean EUR~53,723.79, median EUR~0.00, 99th percentile EUR~123,864.08, and maximum EUR~1,399,852,194.90. The incoming (creditor-side) 30-day maturity target is even sparser, equaling zero in 97.17\% of complete rows.

\begin{figure}[tbp]
  \centering
  \includegraphics[width=\textwidth]{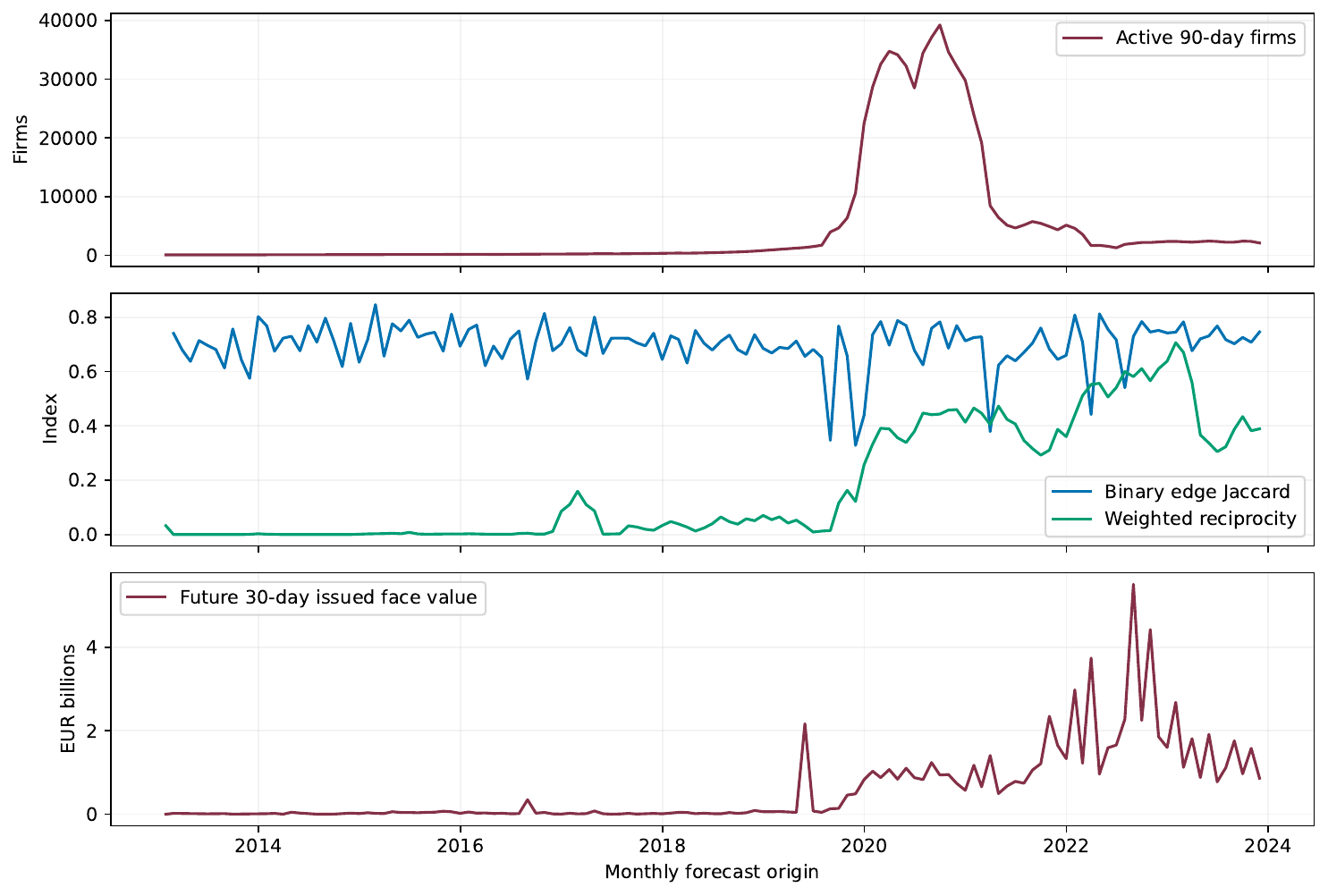}
  \caption{Monthly evolution of active 90-day firms, binary edge Jaccard persistence, weighted reciprocity, and aggregate future 30-day issued face value across forecast origins (2013--2023).}
  \label{fig:network-evolution}
\end{figure}

Network scale and structural continuity shift markedly across the decade (Figure~\ref{fig:network-evolution}). During the 2019--2023 held-out period, annual firm-origin counts swing from 56,872 (2019) to 398,598 (2020), 98,537 (2021), 27,236 (2022), and 25,398 (2023). Concurrently, annual primary zero-target shares range from 74.58\% (2019) to 90.96\% (2021), dual-role shares rise from 4.37\% (2020) to 50.93\% (2022), and prior-year static embedding availability spans 13.36\% (2019) to 83.96\% (2021). Consequently, static-versus-rolling comparisons inherently reflect missing-node coverage and curation regime shifts alongside algorithmic recency.

\subsection{Primary Model Benchmark}
\label{sec:res-primary}

\begin{figure}[tbp]
  \centering
  \includegraphics[width=\textwidth]{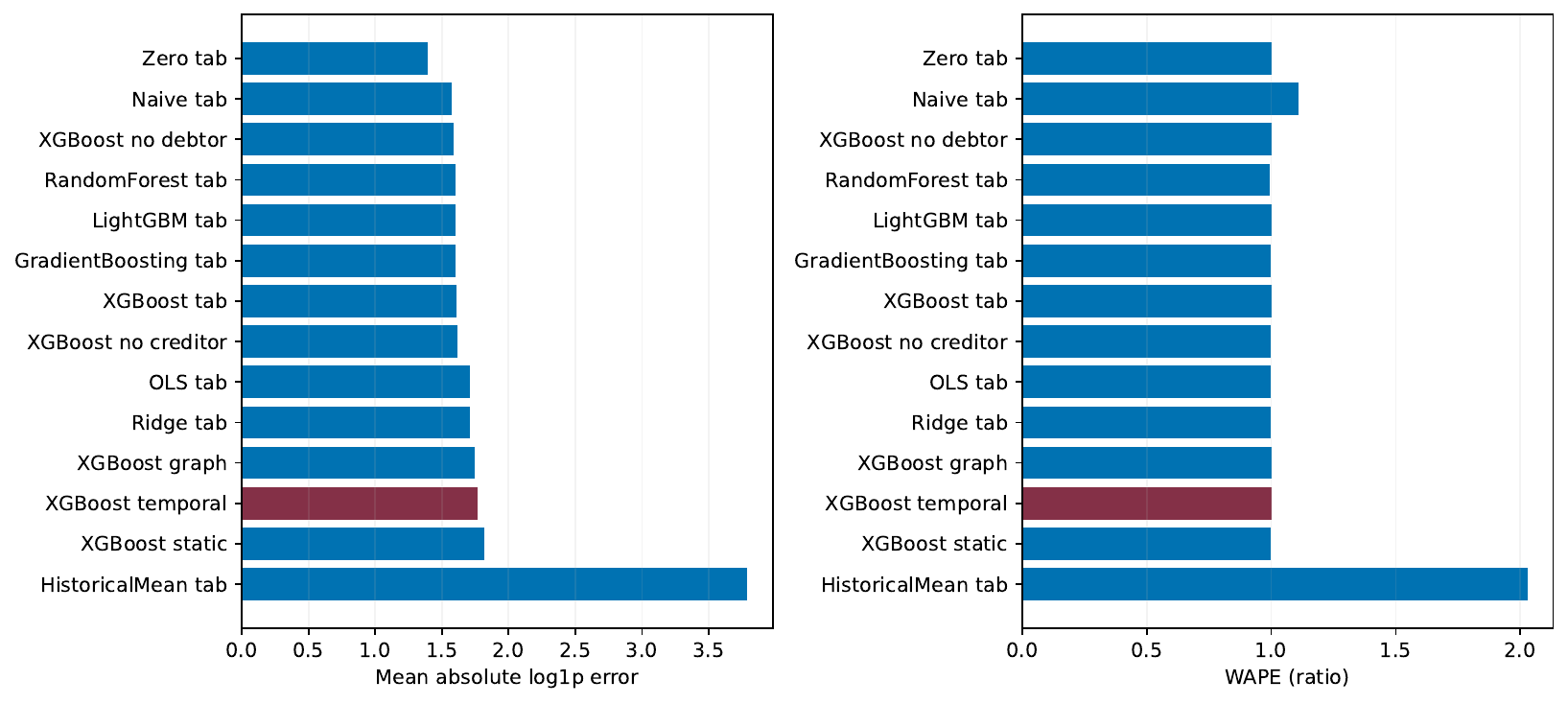}
  \caption{Primary outgoing 30-day maturity forecast accuracy on two scales: pooled $\log(1+\mathrm{EUR})$ MAE (left) and WAPE ratio (right), ordered by log-MAE.}
  \label{fig:model-comparison}
\end{figure}

Table~\ref{tab:primary-results} reports pooled performance across all 606,641 complete held-out firm-origin observations (59 monthly dates) for the primary outgoing 30-day new-maturity task, and Figure~\ref{fig:model-comparison} contrasts pooled log-MAE against WAPE.

\begin{table}[tbp]
  \centering
  \caption{Pooled held-out performance (2019--2023; 606,641 firm-origin rows across 59 monthly dates) for primary outgoing 30-day new-maturity exposure. Coverage denotes empirical containment of nominal 90\% split-residual intervals.}
  \label{tab:primary-results}
  \begin{adjustbox}{max width=\textwidth}
  \begin{tabular}{@{}lrrrrrr@{}}
    \toprule
    \textbf{Model} & \textbf{Log-MAE} & \textbf{MAE (EUR)} & \textbf{RMSE (EUR)} & \textbf{WAPE (\%)} & \textbf{$R^2$} & \textbf{Coverage (\%)} \\
    \midrule
    Zero baseline & \textbf{1.391} & 53,800 & $4.24 \times 10^6$ & 100.0 & $-0.000161$ & 91.1 \\
    30-day naive rate & 1.571 & 59,600 & $\mathbf{4.01 \times 10^6}$ & 110.8 & \textbf{0.106000} & 86.2 \\
    365-day historical rate & 3.791 & 109,000 & $4.79 \times 10^6$ & 203.0 & $-0.277000$ & 87.7 \\
    OLS (tabular) & 1.708 & 53,600 & $4.24 \times 10^6$ & 99.8 & 0.000676 & 90.7 \\
    Ridge (tabular) & 1.708 & 53,600 & $4.24 \times 10^6$ & 99.8 & 0.000621 & 90.7 \\
    Random Forest (tabular) & 1.599 & \textbf{53,500} & $4.24 \times 10^6$ & \textbf{99.6} & 0.000334 & 90.2 \\
    Histogram Boosting (tabular) & 1.603 & 53,600 & $4.24 \times 10^6$ & \textbf{99.6} & 0.000964 & 90.6 \\
    XGBoost (tabular) & 1.606 & 53,800 & $4.24 \times 10^6$ & 100.0 & 0.001450 & 91.1 \\
    LightGBM (tabular) & 1.600 & 53,700 & $4.24 \times 10^6$ & 100.0 & 0.001180 & 90.6 \\
    XGBoost + directed graph & 1.749 & 53,800 & $4.24 \times 10^6$ & 100.0 & 0.001700 & 91.5 \\
    XGBoost + static embeddings & 1.821 & 53,700 & $4.23 \times 10^6$ & 99.9 & 0.003040 & 91.8 \\
    XGBoost + rolling embeddings & 1.767 & 53,800 & $4.23 \times 10^6$ & 100.0 & 0.003370 & 91.7 \\
    XGBoost (no-debtor ablation) & 1.588 & 53,900 & $4.23 \times 10^6$ & 100.2 & 0.006850 & 91.1 \\
    XGBoost (no-creditor ablation) & 1.615 & 53,600 & $4.22 \times 10^6$ & 99.8 & 0.008790 & 90.6 \\
    \bottomrule
  \end{tabular}
  \end{adjustbox}
\end{table}

\begin{figure}[tbp]
  \centering
  \includegraphics[width=\textwidth]{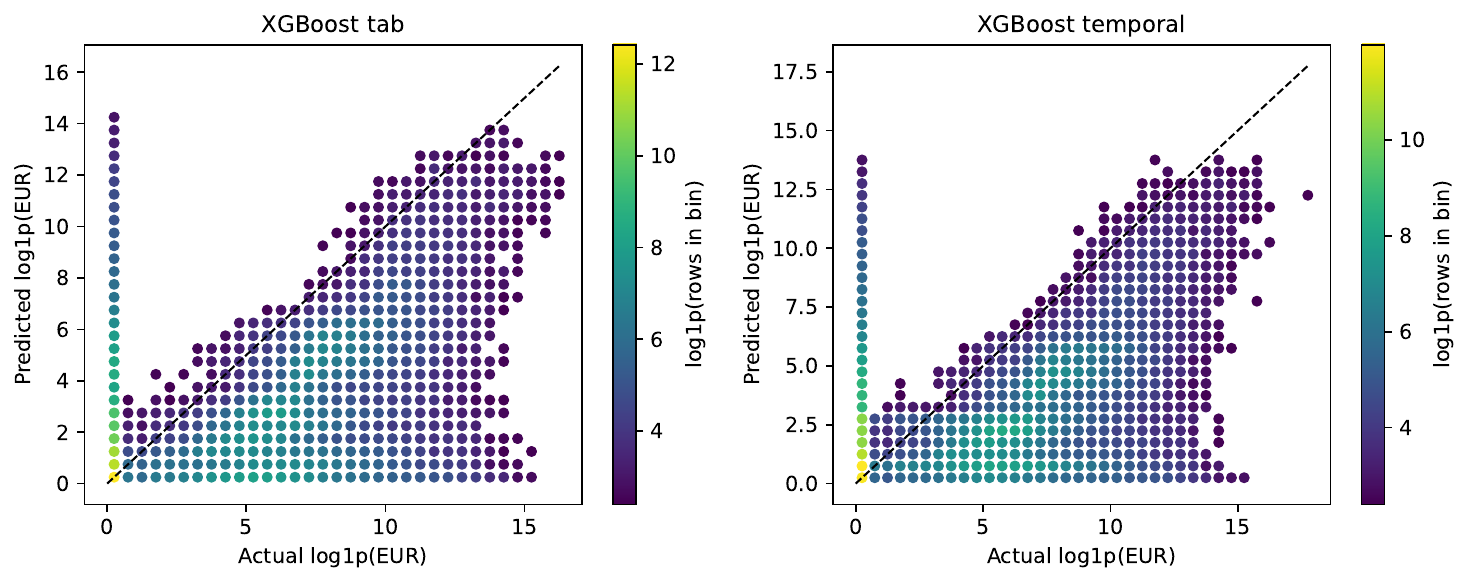}
  \caption{Binned actual-versus-predicted density for primary outgoing 30-day maturity exposure comparing tabular XGBoost (left) and rolling XGBoost (right) in 0.5-wide $\log(1+\mathrm{EUR})$ bins (bins with $<10$ rows suppressed for privacy; dashed line indicates equality).}
  \label{fig:actual-prediction}
\end{figure}

The empirical pattern is unambiguous. Under pooled log-MAE, the constant zero baseline achieves the lowest error (1.391) because 79.28\% of panel outcomes (and up to 90.96\% in 2021) are exactly zero, penalizing learned squared-error models that predict small positive conditional log expectations. Among learned models, tabular tree ensembles cluster tightly between 1.599 (Random Forest) and 1.606 (XGBoost tabular), whereas adding directed graph features raises pooled log-MAE to 1.749, appending rolling embeddings raises it further to 1.767 (exact raw MAE EUR~53,773.64; RMSE EUR~4,233,246.95; WAPE 100.04\%; $R^2 = 0.00337$), and appending static embeddings yields 1.821. On the raw EUR scale, all learned models cluster near 100\% WAPE and near-zero $R^2$ due to extreme tail invoices, while the 30-day naive rate achieves the highest raw $R^2$ (0.106) at the cost of elevated WAPE (110.8\%). Figure~\ref{fig:actual-prediction} illustrates strong central compression and substantial dispersion relative to the equality line for both tabular and rolling XGBoost.

\subsection{Prespecified Hypothesis Contrasts (H1--H3)}
\label{sec:res-hypotheses}

Table~\ref{tab:hypothesis-contrasts} presents the equal-date circular six-month block bootstrap tests controlling the primary three-test family via Holm's procedure.

\begin{table}[tbp]
  \centering
  \caption{Prespecified equal-date primary log-MAE contrasts (rolling XGBoost minus comparator) across 59 monthly origins using 999 circular six-month block bootstrap draws and Holm adjustment.}
  \label{tab:hypothesis-contrasts}
  \begin{adjustbox}{max width=\textwidth}
  \begin{tabular}{@{}llrrrr@{}}
    \toprule
    \textbf{Hypothesis} & \textbf{Comparator} & \textbf{Mean Difference} & \textbf{95\% Block CI} & \textbf{Raw $p$} & \textbf{Holm $p$} \\
    \midrule
    H1 & XGBoost tabular & $+0.05599$ & $[-0.01438,\; +0.13236]$ & 0.131 & 0.262 \\
    H2 & XGBoost + directed graph & $+0.01823$ & $[+0.00358,\; +0.03031]$ & 0.009 & \textbf{0.027} \\
    H3 & XGBoost + static embeddings & $-0.00699$ & $[-0.03725,\; +0.01969]$ & 0.616 & 0.616 \\
    \bottomrule
  \end{tabular}
  \end{adjustbox}
\end{table}

None of H1--H3 supports an advantage for rolling temporal embeddings. For H1, the equal-date temporal-minus-tabular contrast is $+0.05599$ log units (95\% CI $[-0.01438, 0.13236]$; Holm-adjusted $p = 0.262$), leaving the contrast statistically inconclusive with a point estimate favoring tabular XGBoost. For H2, appending rolling embeddings to the directed graph-feature model produces a statistically significant degradation of $+0.01823$ log units (95\% CI $[0.00358, 0.03031]$; Holm-adjusted $p = 0.027$). For H3, rolling embeddings do not differ detectably from prior-year static embeddings ($-0.00699$; 95\% CI $[-0.03725, 0.01969]$; Holm-adjusted $p = 0.616$).

\subsection{Supplementary Firm-Level and Network-Level Benchmarks}
\label{sec:res-supp}

Tables~\ref{tab:supplementary-firm} and~\ref{tab:network-totals} summarize the lowest-error tested configurations across all twelve firm-level tasks (spanning count, face value, and new maturity across debtor/creditor roles and 30/60/90-day horizons) and all nine aggregate network-level tasks. Across supplementary firm tasks, simple 30-day naive rates, tabular Random Forests, tabular XGBoost, and zero baselines consistently achieve the lowest errors; neither directed graph features nor rolling embeddings win any supplementary task on either scale. At the aggregate monthly network level (57--59 test origins), 30-day naive rates, 365-day historical rates, and Ridge regression dominate across issuance count, face value, and maturity-weighted exposure.

\begin{table}[tbp]
  \centering
  \caption{Descriptive lowest-error configurations among executed firm-level tasks (2019--2023). Counts are in invoice units; amount and maturity targets are in EUR.}
  \label{tab:supplementary-firm}
  \begin{adjustbox}{max width=\textwidth}
  \begin{tabular}{@{}llrlrlr@{}}
    \toprule
    \textbf{Target} & \textbf{Role} & \textbf{Days} & \textbf{Lowest Log-MAE Model} & \textbf{Log-MAE} & \textbf{Lowest Raw-MAE Model} & \textbf{Raw MAE} \\
    \midrule
    amount & in & 30 & 30-day naive rate & 0.627 & Random Forest & 109,000 \\
    amount & out & 30 & Random Forest & 2.243 & Random Forest & 110,000 \\
    amount & out & 60 & 30-day naive rate & 2.368 & 365-day historical rate & 208,000 \\
    amount & out & 90 & 30-day naive rate & 2.521 & 365-day historical rate & 309,000 \\
    count & in & 30 & 30-day naive rate & 0.074 & 30-day naive rate & 0.581 \\
    count & out & 30 & Random Forest & 0.300 & OLS & 0.793 \\
    count & out & 60 & XGBoost tabular & 0.406 & XGBoost tabular & 1.440 \\
    count & out & 90 & XGBoost tabular & 0.492 & XGBoost tabular & 2.060 \\
    mature & in & 30 & 30-day naive rate & 0.275 & Histogram Boosting & 54,000 \\
    mature & out & 30 & Zero baseline & 1.391 & Random Forest & 53,500 \\
    mature & out & 60 & 30-day naive rate & 2.273 & 30-day naive rate & 136,000 \\
    mature & out & 90 & XGBoost tabular & 2.677 & 30-day naive rate & 254,000 \\
    \bottomrule
  \end{tabular}
  \end{adjustbox}
\end{table}

\begin{table}[tbp]
  \centering
  \caption{Descriptive lowest-error configurations among executed aggregate network-level forecasting tasks (2019--2023). Each invoice enters network totals once.}
  \label{tab:network-totals}
  \begin{adjustbox}{max width=\textwidth}
  \begin{tabular}{@{}lrlrlrr@{}}
    \toprule
    \textbf{Network Target} & \textbf{Days} & \textbf{Lowest Log-MAE Model} & \textbf{Log-MAE} & \textbf{Lowest Raw-MAE Model} & \textbf{Raw MAE} & \textbf{Dates} \\
    \midrule
    amount & 30 & 30-day naive rate & 0.519 & Ridge & $6.49 \times 10^8$ & 59 \\
    amount & 60 & 30-day naive rate & 0.490 & 30-day naive rate & $1.18 \times 10^9$ & 58 \\
    amount & 90 & 30-day naive rate & 0.517 & 30-day naive rate & $1.78 \times 10^9$ & 57 \\
    count & 30 & 30-day naive rate & 0.264 & 30-day naive rate & 2,560 & 59 \\
    count & 60 & 30-day naive rate & 0.251 & 30-day naive rate & 5,670 & 58 \\
    count & 90 & 30-day naive rate & 0.276 & 30-day naive rate & 9,630 & 57 \\
    weighted & 30 & 30-day naive rate & 0.608 & Ridge & $1.96 \times 10^8$ & 59 \\
    weighted & 60 & 30-day naive rate & 0.559 & 365-day historical rate & $3.38 \times 10^8$ & 58 \\
    weighted & 90 & 365-day historical rate & 0.580 & 365-day historical rate & $3.47 \times 10^8$ & 57 \\
    \bottomrule
  \end{tabular}
  \end{adjustbox}
\end{table}

\section{Explainability Results and Attribution Stability}
\label{sec:xai-results}

\begin{figure}[tbp]
  \centering
  \includegraphics[width=\textwidth]{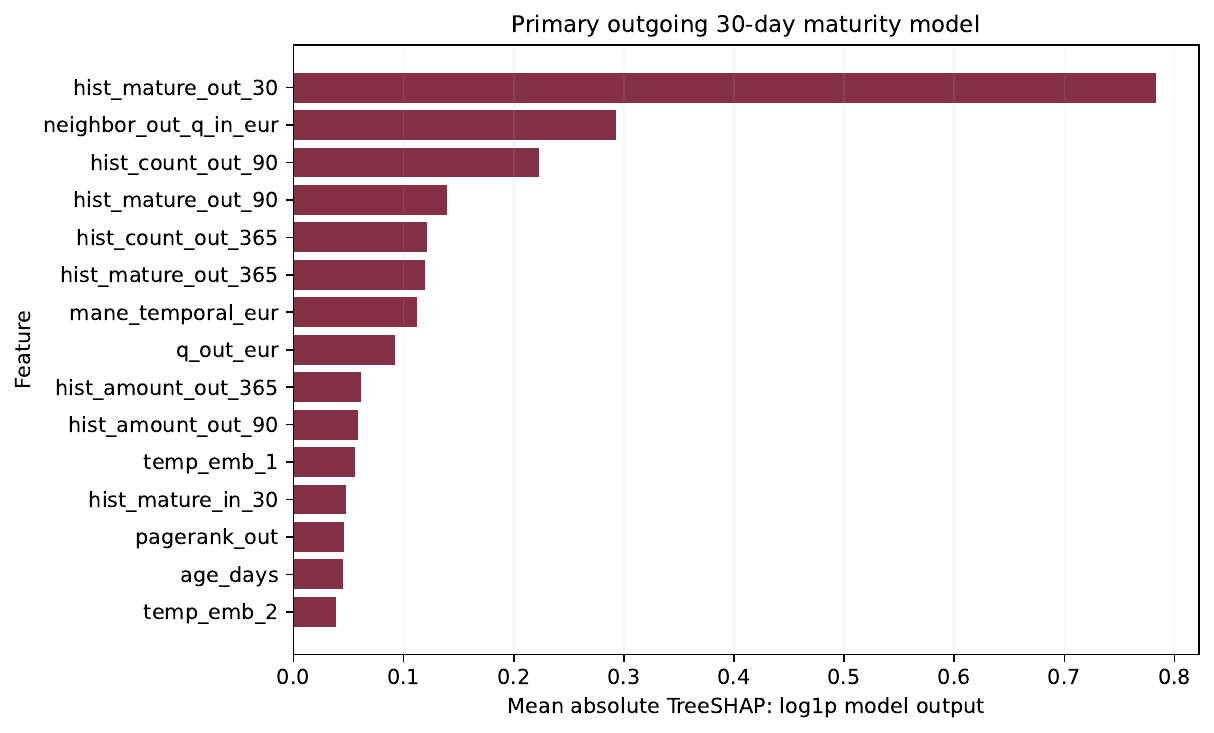}
  \caption{Global TreeSHAP feature importance (mean absolute attribution on raw $\log(1+\mathrm{EUR})$ output) for the primary outgoing 30-day maturity rolling XGBoost model.}
  \label{fig:shap-global}
\end{figure}

Across 2,500 held-out observations (500 per year across 2019--2023), interventional TreeSHAP verifies local additivity with maximum residual $6.32 \times 10^{-6}$ raw log-output units. Addressing H4, global absolute-attribution shares separate sharply across feature families: own histories and booked maturities account for 67.24\% of total absolute attribution, directed graph and neighbor features contribute 23.51\%, and aligned rolling embeddings account for 9.24\%. As shown in Figure~\ref{fig:shap-global}, the top five individual predictors are \code{hist_mature_out_30}, \code{neighbor_out_q_in_eur}, \code{hist_count_out_90}, \code{hist_mature_out_90}, and \code{hist_count_out_365}, followed by \code{hist_mature_out_365}, \code{mane_temporal_eur}, and \code{q_out_eur}.

\begin{figure}[tbp]
  \centering
  \includegraphics[width=\textwidth]{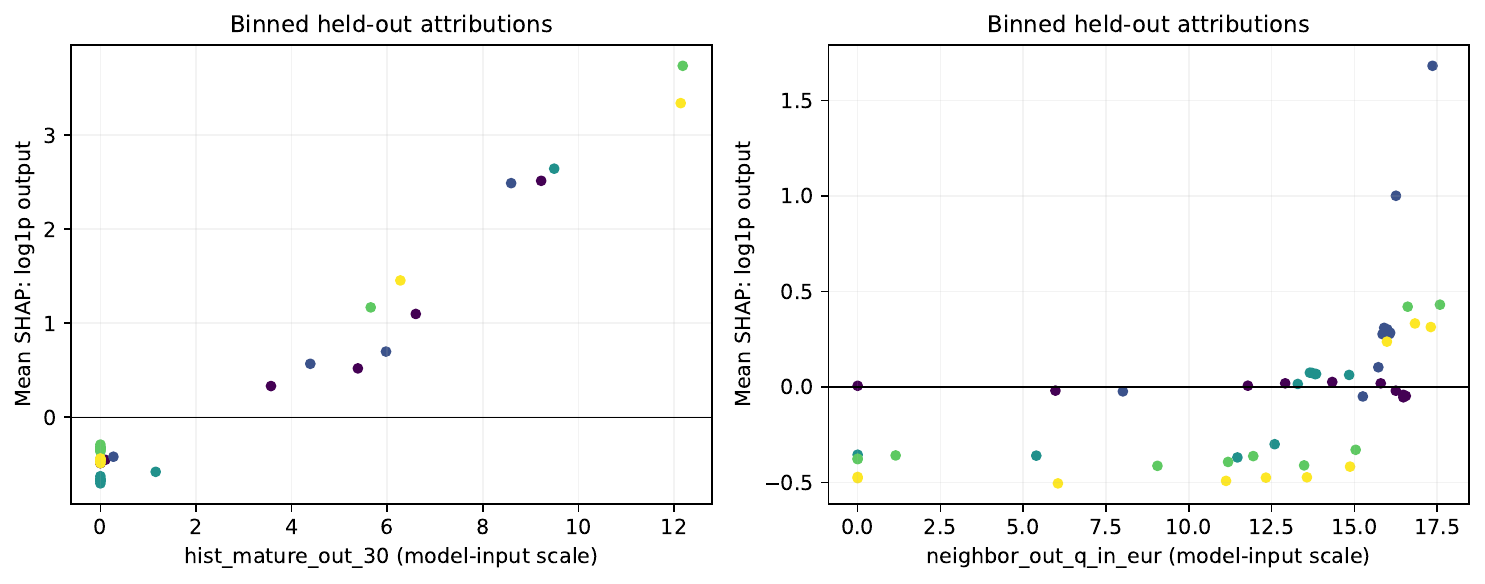}
  \caption{Binned held-out TreeSHAP dependence profiles (10 bins per year) for the two highest-attribution predictors: \texttt{hist\_mature\_out\_30} (left) and \texttt{neighbor\_out\_q\_in\_eur} (right).}
  \label{fig:shap-dependence}
\end{figure}

Figure~\ref{fig:shap-dependence} plots binned held-out TreeSHAP dependence profiles for the top two features (\code{hist_mature_out_30} and \code{neighbor_out_q_in_eur}). While trailing 30-day outgoing maturity exhibits a clear monotonic positive relationship with raw log-scale model predictions across all five years, the neighbor proximity feature displays visibly greater cross-year dispersion. Across the full feature vector, pairwise year-to-year Spearman rank correlations of mean absolute TreeSHAP range from 0.486 (2020 vs.\ 2023) to 0.651 (2019 vs.\ 2020), indicating moderate rank persistence alongside non-trivial year-to-year redistribution (Figure~\ref{fig:shap-stability}; complete annual family shares and rank-correlation matrices appear in Appendix~\ref{app:supplementary}). Crucially, the fact that directed graph features and embeddings receive 32.75\% combined prediction attribution while failing to improve out-of-sample log-MAE illustrates the distinction between prediction attribution and accuracy contribution emphasized by \citet{hue2026measuring}.

\begin{figure}[tbp]
  \centering
  \includegraphics[width=\textwidth]{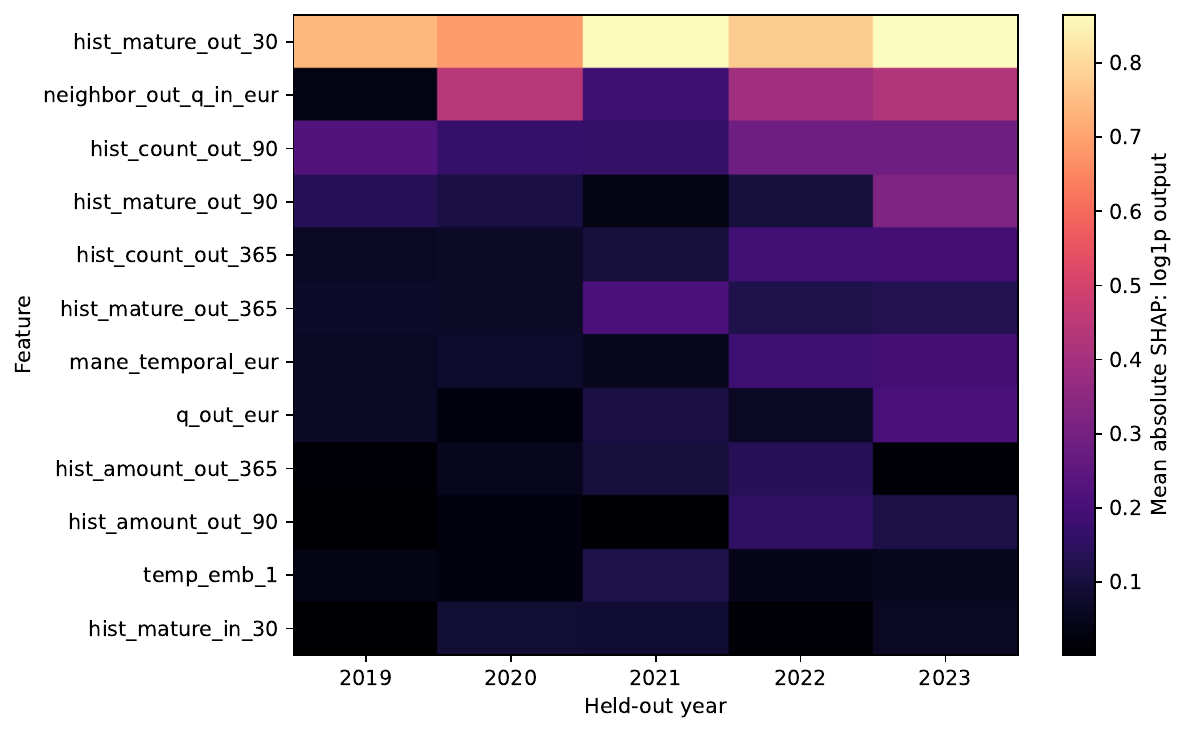}
  \caption{Cross-year TreeSHAP attribution stability across 2019--2023 for the twelve highest-importance predictors on a common color scale.}
  \label{fig:shap-stability}
\end{figure}

\section{Robustness, Temporal Generalization, and Computational Cost}
\label{sec:robustness}

\begin{figure}[tbp]
  \centering
  \includegraphics[width=\textwidth]{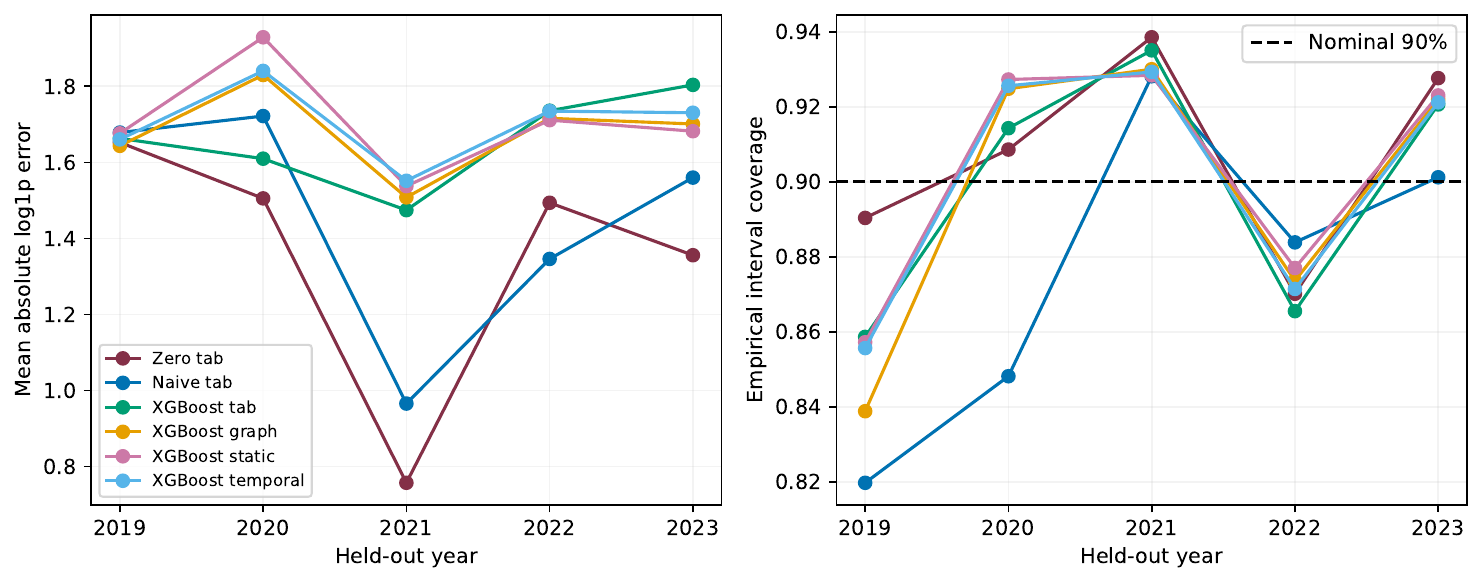}
  \caption{Held-out annual primary log-MAE (left) and empirical containment of nominal 90\% split-residual intervals (right) across 2019--2023.}
  \label{fig:year-errors-coverage}
\end{figure}

\begin{figure}[tbp]
  \centering
  \includegraphics[width=\textwidth]{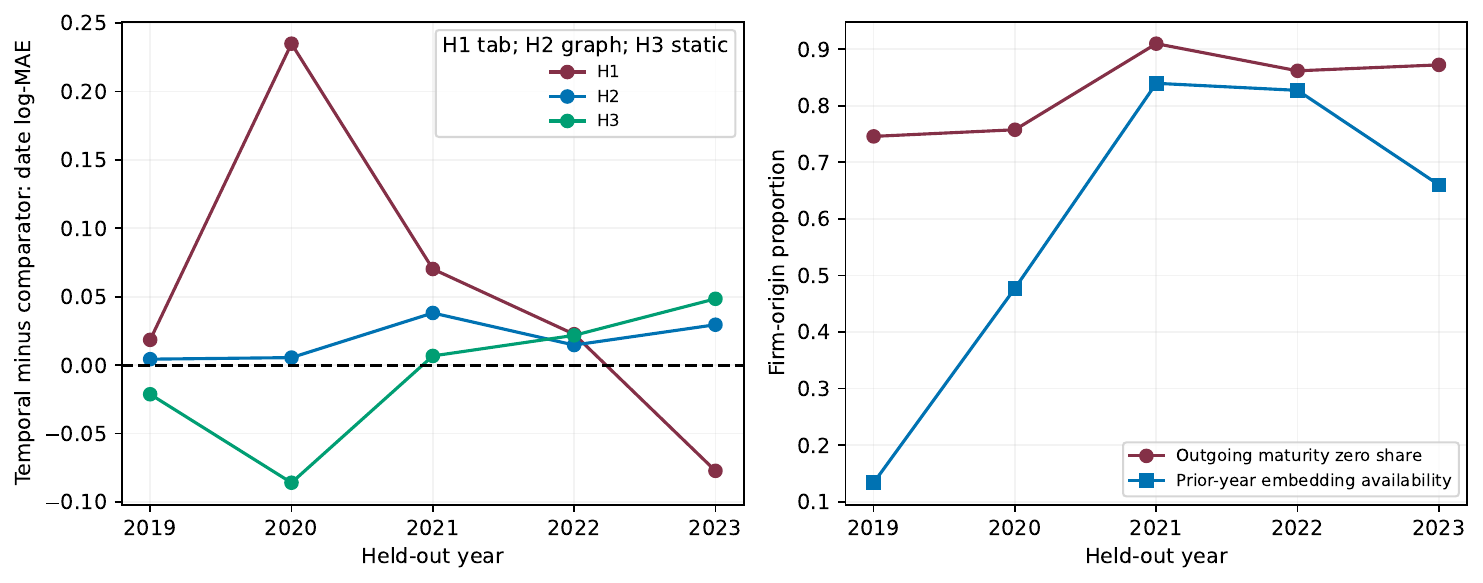}
  \caption{Annual equal-date log-MAE contrasts for H1--H3 (left; negative favors rolling) alongside annual outgoing maturity zero share and prior-year static embedding availability (right).}
  \label{fig:temporal-generalisation}
\end{figure}

Figure~\ref{fig:year-errors-coverage} tracks annual held-out log-MAE and nominal 90\% interval coverage across 2019--2023. Rolling XGBoost empirical annual coverage ranges from 85.57\% to 92.93\% (pooled 91.70\%), reflecting unadapted pre-test calibration intervals facing shifting yearly zero shares and tail magnitudes. Figure~\ref{fig:temporal-generalisation} disaggregates equal-date annual H1--H3 contrasts alongside annual zero-target shares and prior-year embedding availability. While H1 favors tabular XGBoost in 2019--2022 (peaking at $+0.235$ in 2020) and turns negative only in 2023 ($-0.077$), H2 remains positive (favoring directed graph features over rolling embeddings) in every single held-out year ($+0.004$ to $+0.038$).

Block-length and metric-scale sensitivity checks confirm robustness of the null/negative finding (Appendix~\ref{app:supplementary}, Table~\ref{tab:block-sensitivity}). Across 3-, 6-, and 12-month circular blocks, H1 remains inconclusive (unadjusted $p \in [0.094, 0.183]$), H2 remains significantly adverse for rolling embeddings ($p \in [0.005, 0.014]$), and H3 remains indistinguishable from zero ($p \in [0.506, 0.704]$). On the raw-EUR MAE scale with 6-month blocks, the equal-date H1 contrast is EUR~$+8.87$ (95\% CI $[\text{EUR}\,-1{,}335.43,\; \text{EUR}\,1{,}542.24]$, $p = 0.995$), confirming zero economically meaningful gain.

\begin{figure}[tbp]
  \centering
  \includegraphics[width=\textwidth]{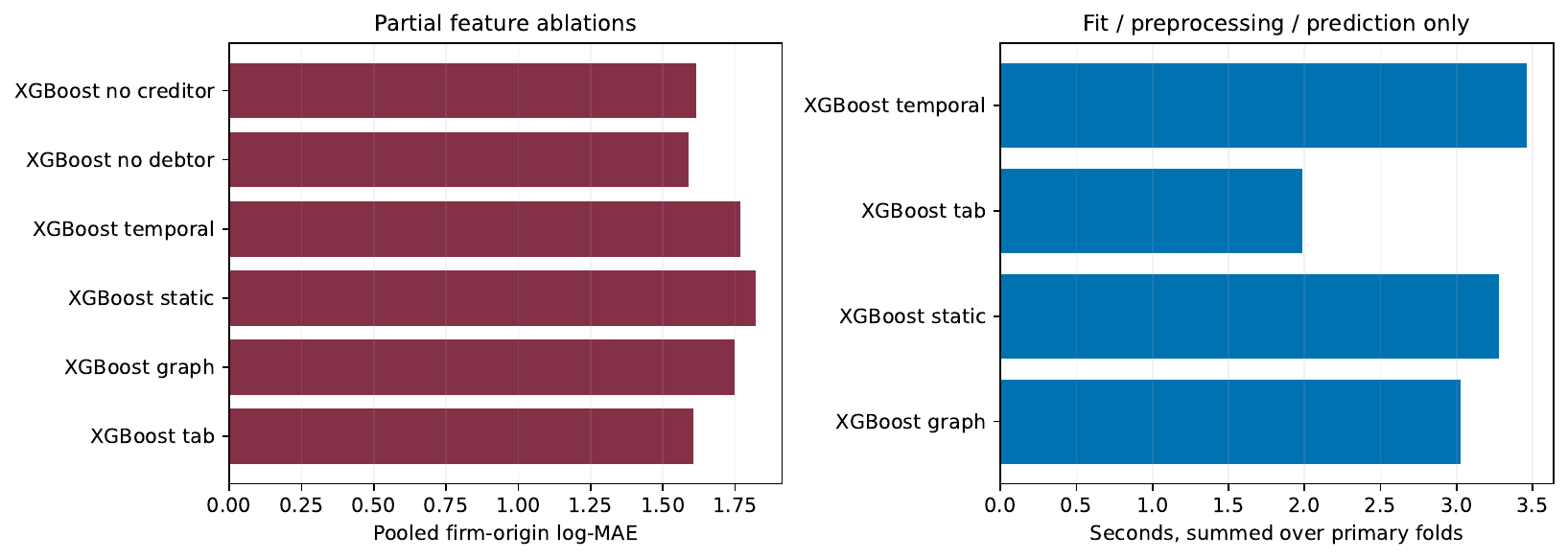}
  \caption{Partial directional feature ablations (left) and cumulative primary-fold preprocessing, fitting, and prediction runtimes across XGBoost feature blocks (right).}
  \label{fig:ablation-efficiency}
\end{figure}

Partial directional ablations and runtime costs appear in Figure~\ref{fig:ablation-efficiency}. Removing outgoing-suffix graph columns (\code{no_debtor}) lowers pooled log-MAE from 1.767 to 1.588 (below tabular XGBoost at 1.606), whereas removing incoming-suffix graph columns (\code{no_creditor}) yields 1.615; because own histories, shared-role features, and undirected embeddings remain present, these results indicate sensitivity to noisy high-dimensional directional graph covariates under the 5,000-row training cap rather than total irrelevance of bilateral roles. Summed primary-fold preprocessing, fitting, and prediction time rises from 1.99~s (tabular XGBoost) to 3.03~s (directed graph), 3.28~s (static), and 3.46~s (rolling), excluding upstream graph construction and random-walk embedding overhead.

\section{Discussion}
\label{sec:discussion}

The empirical evidence rejects the assumption that appending temporal graph embeddings automatically improves firm-level contractual invoice forecasting. Across 606,641 held-out firm-origin observations, rolling \texttt{node2vec} XGBoost records higher pooled log-MAE (1.767) than tabular XGBoost (1.606), fails to establish an equal-date gain under H1 (Holm $p = 0.262$), performs significantly worse than explicit directed graph features under H2 (Holm $p = 0.027$), and shows no detectable advantage over prior-year static embeddings under H3 (Holm $p = 0.616$). Equally striking, the constant zero baseline attains the lowest pooled log-MAE (1.391), underscoring why zero-heavy financial regression benchmarks must retain trivial baselines to prevent overclaiming modest improvements over weak comparators \citep{huang2023temporal,kureljusic2023applicability}.

Target framing explains much of this behavior. Separating cutoff-known booked maturities $B_{v,t,h}^r$ from newly issued within-horizon maturities $Y_{v,t,h}^{\mathrm{mature}, r}$ removes predictable deterministic backlog from the target and exposes the genuine sparsity of short-horizon new issuance (79.28\% zeros overall; 97.17\% on the creditor side). Under squared-error training on $\log(1+Y)$ with a 5,000-row per-fold fitting cap, adding noisy undirected embedding coordinates increases variance and shifts predictions slightly above zero on inactive firm-origins, inflating log-MAE while delivering negligible raw-EUR variance explanation ($R^2 = 0.0034$).

Relative to our companion study \citep{madugula2026beyond}, this manuscript shifts cleanly from retrospective structural analysis and contemporaneous concentration modeling to explicit cutoff-safe future labeling, multi-learner out-of-sample benchmarking, and explanation stability auditing. Reusing the cent-preserving graph engine and MANE indicator as candidate predictors demonstrates that structurally informative systemic exposure indicators do not automatically translate into superior firm-level short-horizon predictive accuracy. Likewise, contrasting our ledger with externally labeled banking studies \citep{berloco2021predicting,cheng2020delinquent} clarifies why observed payment and default data remain indispensable when moving from contractual issuance monitoring to credit-loss or liquidity-contagion forecasting.

\section{Theoretical and Practical Implications}
\label{sec:implications}

Methodologically, this benchmark formalizes a clean separation among three distinct contractual quantities: cutoff-known booked maturities, future newly issued face value, and future newly issued within-horizon maturity. Evaluating representation increments against information-rich tabular histories (including booked schedules) and zero/naive baselines enforces empirical discipline across both log-transformed and raw monetary scales. Furthermore, pairing TreeSHAP audits with formal hypothesis contrasts demonstrates empirically that non-trivial SHAP attribution shares (32.75\% combined for graph and embedding predictors) can coexist with zero or negative out-of-sample accuracy gains \citep{hue2026measuring}.

Practically, in authorized working-capital monitoring or supply-chain finance workflows, contractual forecasts should be used solely as triage signals---highlighting portfolios, calendar dates, or counterparties with anticipated issuance bursts or wide residual intervals for human verification. Any actual credit underwriting or invoice-financing decision requires externally verified outstanding balances, legal enforceability, anti-double-financing checks, and counterparty creditworthiness. From a systems-engineering perspective, the accuracy--complexity trade-off strongly favors lean tabular baselines or explicit low-dimensional directed summaries over rolling random-walk embedding pipelines whose construction, alignment, and opacity overheads yield no measured forecasting gain.

\section{Limitations}
\label{sec:limitations}

Several material boundaries govern interpretation of these findings:
\begin{enumerate}
  \item \emph{Resource-bounded training cap:} Although the entire 749,952-invoice ledger is audited and all 606,641 held-out test rows are evaluated intact, each learned training fold is uniformly capped at 5,000 firm-origin rows for memory feasibility. This cap may under-sample rare extreme exposures; conclusions therefore apply to this resource-bounded specification rather than unconstrained full-panel optimization or exhaustive hyperparameter tuning.
  \item \emph{Non-neural temporal representation:} Rolling embeddings use undirected weighted \texttt{node2vec} with past-only orthogonal Procrustes alignment rather than end-to-end continuous-time neural architectures (such as TGN or DySAT). Undirected projection discards edge orientation inside the 8-dimensional embedding block.
  \item \emph{Ledger provenance and unobserved outcomes:} Payments, defaults, recoveries, firm sizes, industry codes, and operational record-publication timestamps are unobserved. The separately curated 2021 corpus lacks a cross-year entity crosswalk, and 13 extreme literal due dates remain unaltered in the analytical panel.
  \item \emph{Statistical and explanation scope:} Fixed split-residual intervals yield empirical coverage only without distribution-free nonstationary guarantees. TreeSHAP evaluates a single primary learner across 2,500 sampled held-out rows and does not identify causal effects or span network-total models.
\end{enumerate}

\section{Conclusion}
\label{sec:conclusion}

Within a strictly leakage-audited, horizon-purged, resource-bounded forecasting benchmark spanning 749,952 real-world B2B invoices (2012--2023) and 606,641 held-out firm-origin observations (2019--2023), temporal graph embeddings do not improve forecasts of newly issued 30-day debtor-side contractual maturity exposure. Hypothesis H1 finds no statistically detectable improvement over tabular XGBoost (Holm-adjusted $p = 0.262$), H2 shows a statistically significant degradation when adding rolling embeddings to directed graph features (Holm-adjusted $p = 0.027$), and H3 detects no difference between rolling and prior-year static embeddings (Holm-adjusted $p = 0.616$). Across both firm-level and aggregate network tasks, simple zero, naive-rate, and tabular tree or linear models consistently match or outperform more complex graph-embedding pipelines. Held-out TreeSHAP confirms that own-firm histories and booked maturities drive 67.24\% of raw log-output attribution. Rather than proving that temporal graphs can never aid receivables forecasting, this study establishes a transparent, reproducible negative baseline demonstrating that representation complexity must earn its place against cutoff-safe tabular and trivial benchmarks.

\section*{Data Availability, Ethics, and Declarations}
\addcontentsline{toc}{section}{Data Availability, Ethics, and Declarations}

\noindent\textbf{Data and Code Availability.} The empirical analysis uses the pseudonymized atomic invoice ledger archived on Mendeley Data under a CC BY 4.0 license \citep{esteva2026dataset} (DOI: \href{https://doi.org/10.17632/28rbmvwsm9.1}{\nolinkurl{10.17632/28rbmvwsm9.1}}; SHA-256 master digest \texttt{\seqsplit{e139b8dc75b7334d243874b6ebdaf1798c9969038b193f7eb4cd0e873bcc1a88}}), with companion curation utilities at \url{https://github.com/peplluis7/onemillioninvoices} and prior structural analysis reported in \citet{madugula2026beyond}. All executed Python scripts (\code{invoice_graph_core.py}, \code{run_forecasting_benchmark.py}, \code{run_selftest.py}, \code{run_diagnostics.py}, \code{build_reporting_tables.py}, \code{replot_empirical_figures.py}, \code{render_conceptual_diagrams.py}), pinned dependency specifications, six Mermaid diagram specifications, raster copies of all 16 figures, and 25 privacy-safe aggregate CSV/JSON outputs are bundled in the arXiv ancillary directory (\code{anc/}). Raw invoices, firm-origin panels, individual forecasts, and individual SHAP vectors are excluded to preserve transaction-level confidentiality.

\medskip
\noindent\textbf{Ethics and Privacy Statement.} All ledger records were pseudonymized prior to public release; no re-identification attempt was performed, no human-subjects intervention occurred, and all released visual and tabular outputs are aggregated or bin-censored (minimum 10 observations per bin).

\medskip
\noindent\textbf{Declaration of Generative AI and AI-Assisted Technologies in the Writing and Analytical Process.} In accordance with publisher fair-use and research-integrity policies, generative AI assistance was used under author supervision for Python code scaffolding, LaTeX formatting, manuscript copy-editing, and bibliographic verification checks. All empirical computations, automated leakage/metric tests, scientific interpretations, and final text were independently audited and verified by the authors, who take full responsibility for the integrity and accuracy of the published work.

\medskip
\noindent\textbf{CRediT Author Contributions.} \emph{Sai Srikanth Madugula:} Conceptualization, Methodology, Software, Validation, Formal analysis, Visualization, Writing -- original draft, Writing -- review \& editing. \emph{Peplluis Esteva de la Rosa:} Conceptualization, Data curation, Methodology, Supervision, Writing -- review \& editing. \emph{Daya Shankar:} Methodology, Validation, Supervision, Writing -- review \& editing.

\medskip
\noindent\textbf{Funding.} This research received no specific grant from any funding agency in the public, commercial, or not-for-profit sectors.

\medskip
\noindent\textbf{Declaration of Competing Interest.} The authors declare that they have no known competing financial interests or personal relationships that could have appeared to influence the work reported in this paper.

\appendix

\section{Original Data Dictionary and Variable Operationalization}
\label{app:dictionary}

Table~\ref{tab:data-dictionary} records the exact eight-column schema of the master ledger. Complete per-variable operationalization across all firm-panel and network-panel predictors, targets, transformations, and missingness rules is archived in \code{anc/tables/variable_operationalisation.csv}.

\begin{table}[tbp]
  \centering
  \caption{Observed eight-column schema of the audited master invoice ledger (\texttt{original\_data\_dictionary.csv}).}
  \label{tab:data-dictionary}
  \begin{adjustbox}{max width=\textwidth}
  \begin{tabular}{@{}llllll@{}}
    \toprule
    \textbf{Field} & \textbf{Type} & \textbf{Unit} & \textbf{Definition} & \textbf{Preservation} & \textbf{Missingness Rule} \\
    \midrule
    \code{invoice_code} & String & Pseudonym & Unique recorded invoice ID & Source preserved & Validation halts if missing \\
    \code{debtor_code} & String & Pseudonym & Recorded owing firm ID & Source preserved & Validation halts if missing \\
    \code{creditor_code} & String & Pseudonym & Recorded claiming firm ID & Source preserved & Validation halts if missing \\
    \code{amount_cents} & Integer & EUR cents & Literal recorded face value & Source preserved & Positive integer required \\
    \code{issue_date} & ISO string & Calendar date & Literal recorded issue date & Source preserved & Valid ISO date required \\
    \code{due_date} & ISO string & Calendar date & Literal contractual due date & Source preserved & Valid date with $u_i \ge s_i$ \\
    \code{issue_cohort} & Integer & Year & Recorded issue calendar year & Source preserved & Must match \code{issue_date} \\
    \code{analysis_role} & String & Category & \code{issue_cohort} or \code{terminal_bridge_only} & Source preserved & Strict role partitioning \\
    \bottomrule
  \end{tabular}
  \end{adjustbox}
\end{table}

\section{Target Filters, Cutoff Conventions, and Observation Shifts}
\label{app:targets}

All monthly origins span 31 January 2013 through 30 November 2023 (131 origins). Active-firm eligibility at origin $t$ requires at least one issued invoice in either role during $(t-90, t]$. Incomplete terminal horizons ($t + h > \text{2023-12-31}$) are masked as \code{NaN}. Table~\ref{tab:year-shift} documents the pronounced annual compositional and sparsity shifts across the five held-out evaluation years.

\begin{table}[tbp]
  \centering
  \caption{Annual observation composition, sparsity, exposure scale, and static-embedding availability across held-out test years (\texttt{year\_shift.csv}).}
  \label{tab:year-shift}
  \begin{adjustbox}{max width=\textwidth}
  \begin{tabular}{@{}rrrrrrrr@{}}
    \toprule
    \textbf{Year} & \textbf{Firm-Origins} & \textbf{Dates} & \textbf{Zero Share (\%)} & \textbf{Mean Out 30d (EUR)} & \textbf{P99 Out 30d (EUR)} & \textbf{Dual-Role (\%)} & \textbf{Static Avail. (\%)} \\
    \midrule
    2019 & 56,872 & 12 & 74.58 & 11,755.92 & 103,308.32 & 6.33 & 13.36 \\
    2020 & 398,598 & 12 & 75.75 & 9,674.51 & 30,594.58 & 4.37 & 47.73 \\
    2021 & 98,537 & 12 & 90.96 & 58,086.50 & 256,275.16 & 15.40 & 83.96 \\
    2022 & 27,236 & 12 & 86.17 & 624,704.81 & 2,880,719.13 & 50.93 & 82.71 \\
    2023 & 25,398 & 11 & 87.22 & 210,416.44 & 1,600,818.26 & 47.08 & 66.01 \\
    \bottomrule
  \end{tabular}
  \end{adjustbox}
\end{table}

\section{Feature Blocks, Preprocessing, and Resource Boundaries}
\label{app:features}

Exact feature memberships for the tabular, directed-graph, static-embedding, rolling-embedding, partial-ablation, and system-level blocks are archived in \code{anc/tables/feature_sets.json}. All scaling, median imputation, and training-row sampling (cap 5,000 rows per eligible fold, seed $1729 + Y_{\mathrm{test}}$) fit strictly inside each fold's purged training set; full calibration and test panels are evaluated without subsampling.

\section{Leakage Audits and Verification Suite}
\label{app:leakage}

Automated unit and empirical boundary tests executed in \code{anc/run_selftest.py} and \code{anc/run_diagnostics.py} passed all 12 verification checks (\code{selftest.json}) and all 14 independent primary metric recomputations (\code{verification.json}): synthetic target-boundary separation, booked-versus-new maturity isolation, future-node exclusion, prefix feature invariance on truncated ledgers, exact integer-cent aggregation, purged 30/60/90-day split boundaries (\code{leakage_folds.csv}), signed embedding scaling, deterministic walk reproduction, empirical horizon masking, feature--label column disjointness, nonnegative firm ages, and finite aligned embedding coordinates.

\section{Supplementary Sensitivity, SHAP Family, and Stability Tables}
\label{app:supplementary}

Tables~\ref{tab:block-sensitivity}, \ref{tab:shap-family}, and~\ref{tab:shap-stability} report complete block-length and raw-EUR sensitivity contrasts, annual TreeSHAP feature-family magnitudes, and cross-year Spearman rank stability correlations.

\begin{table}[tbp]
  \centering
  \caption{Block-length (3, 6, 12 months) and metric-scale ($\log(1+\mathrm{EUR})$ vs.\ raw EUR MAE) sensitivity for equal-date primary contrasts across 59 held-out origins (\texttt{block\_scale\_sensitivity.csv}).}
  \label{tab:block-sensitivity}
  \begin{adjustbox}{max width=\textwidth}
  \begin{tabular}{@{}llrrrrr@{}}
    \toprule
    \textbf{Contrast} & \textbf{Scale} & \textbf{Block (mo.)} & \textbf{Difference} & \textbf{95\% CI Low} & \textbf{95\% CI High} & \textbf{Raw $p$} \\
    \midrule
    H1 (Rolling $-$ Tabular) & log-MAE & 3 & $+0.05599$ & $-0.00602$ & $+0.12720$ & 0.094 \\
    H1 (Rolling $-$ Tabular) & log-MAE & 6 & $+0.05599$ & $-0.01438$ & $+0.13236$ & 0.131 \\
    H1 (Rolling $-$ Tabular) & log-MAE & 12 & $+0.05599$ & $-0.02504$ & $+0.14469$ & 0.183 \\
    H2 (Rolling $-$ Graph) & log-MAE & 3 & $+0.01823$ & $+0.00394$ & $+0.02979$ & 0.005 \\
    H2 (Rolling $-$ Graph) & log-MAE & 6 & $+0.01823$ & $+0.00358$ & $+0.03031$ & 0.009 \\
    H2 (Rolling $-$ Graph) & log-MAE & 12 & $+0.01823$ & $+0.00302$ & $+0.03004$ & 0.014 \\
    H3 (Rolling $-$ Static) & log-MAE & 3 & $-0.00699$ & $-0.02984$ & $+0.01353$ & 0.506 \\
    H3 (Rolling $-$ Static) & log-MAE & 6 & $-0.00699$ & $-0.03725$ & $+0.01969$ & 0.616 \\
    H3 (Rolling $-$ Static) & log-MAE & 12 & $-0.00699$ & $-0.04293$ & $+0.02511$ & 0.704 \\
    Supp.\ (Graph $-$ Tabular) & log-MAE & 6 & $+0.03776$ & $-0.03861$ & $+0.12298$ & 0.367 \\
    \midrule
    H1 (Rolling $-$ Tabular) & Raw EUR MAE & 6 & $+8.87$ & $-1{,}335.43$ & $+1{,}542.24$ & 0.995 \\
    H2 (Rolling $-$ Graph) & Raw EUR MAE & 6 & $-378.91$ & $-1{,}403.17$ & $+684.23$ & 0.475 \\
    H3 (Rolling $-$ Static) & Raw EUR MAE & 6 & $+207.03$ & $-700.46$ & $+1{,}057.90$ & 0.644 \\
    Supp.\ (Graph $-$ Tabular) & Raw EUR MAE & 6 & $+387.78$ & $-912.12$ & $+1{,}474.24$ & 0.490 \\
    \bottomrule
  \end{tabular}
  \end{adjustbox}
\end{table}

\begin{table}[tbp]
  \centering
  \caption{Annual mean absolute TreeSHAP attributions summed by feature family on raw $\log(1+\mathrm{EUR})$ model outputs (\texttt{shap\_family.csv}).}
  \label{tab:shap-family}
  \begin{adjustbox}{max width=\textwidth}
  \begin{tabular}{@{}rrrrr@{}}
    \toprule
    \textbf{Year} & \textbf{Own History / Booked} & \textbf{Directed Graph / Neighbor} & \textbf{Aligned Rolling Embeddings} & \textbf{Max Additivity Error} \\
    \midrule
    2019 & 1.4247 & 0.2371 & 0.1724 & $5.93 \times 10^{-6}$ \\
    2020 & 1.4912 & 0.7420 & 0.1465 & $2.03 \times 10^{-6}$ \\
    2021 & 1.9749 & 0.5352 & 0.3342 & $3.91 \times 10^{-6}$ \\
    2022 & 2.4007 & 0.9987 & 0.4315 & $6.32 \times 10^{-6}$ \\
    2023 & 2.5761 & 0.9374 & 0.2716 & $5.79 \times 10^{-6}$ \\
    \midrule
    \textbf{Pooled Share} & \textbf{67.24\%} & \textbf{23.51\%} & \textbf{9.24\%} & $\le 6.32 \times 10^{-6}$ \\
    \bottomrule
  \end{tabular}
  \end{adjustbox}
\end{table}

\begin{table}[tbp]
  \centering
  \caption{Pairwise cross-year Spearman rank correlations of mean absolute TreeSHAP feature importance across held-out years (\texttt{shap\_rank\_stability.csv}).}
  \label{tab:shap-stability}
  \begin{adjustbox}{max width=\textwidth}
  \begin{tabular}{@{}lrrrrr@{}}
    \toprule
    \textbf{Year} & \textbf{2019} & \textbf{2020} & \textbf{2021} & \textbf{2022} & \textbf{2023} \\
    \midrule
    2019 & 1.000 & 0.651 & 0.583 & 0.646 & 0.639 \\
    2020 & 0.651 & 1.000 & 0.616 & 0.650 & 0.486 \\
    2021 & 0.583 & 0.616 & 1.000 & 0.566 & 0.553 \\
    2022 & 0.646 & 0.650 & 0.566 & 1.000 & 0.613 \\
    2023 & 0.639 & 0.486 & 0.553 & 0.613 & 1.000 \\
    \bottomrule
  \end{tabular}
  \end{adjustbox}
\end{table}

\section{Executable Software Pipeline and Reproducibility Protocol}
\label{app:pipeline}

Table~\ref{tab:software-index} indexes the complete executable Python modules and conceptual diagram sources archived inside \code{anc/}. Full reproduction commands, exact package versions (Python 3.13, \code{numpy 2.3.5}, \code{pandas 2.2.3}, \code{scipy 1.17.1}, \code{networkx 3.7}, \code{scikit-learn 1.6.1}, \code{gensim 4.4.0}, \code{xgboost-cpu 3.4.1}, \code{lightgbm 4.7.0}, \code{shap 0.54.1}), and environment manifests are documented in \code{anc/README_REPLICATION.txt}, \code{anc/requirements.txt}, and \code{anc/environment.json}.

\begin{table}[tbp]
  \centering
  \caption{Executable replication modules and conceptual diagram assets bundled in the ancillary archive (\texttt{anc/}).}
  \label{tab:software-index}
  \begin{adjustbox}{max width=\textwidth}
  \begin{tabular}{@{}ll@{}}
    \toprule
    \textbf{Ancillary File} & \textbf{Executed Functional Role} \\
    \midrule
    \code{invoice_graph_core.py} & Cent-preserving directed graph engine and schema validator adapted from \citet{madugula2026beyond} \\
    \code{run_forecasting_benchmark.py} & Panel construction, targets, rolling/static \texttt{node2vec}, learner training, intervals, contrasts \\
    \code{run_selftest.py} & 12 synthetic and empirical cutoff, purge, booked-separation, and cent-integrity tests \\
    \code{run_diagnostics.py} & Independent primary metric recomputation, annual shift diagnostics, block sensitivity \\
    \code{build_reporting_tables.py} & Privacy-safe aggregate CSV export, operationalization ledger, binned figure inputs \\
    \code{replot_empirical_figures.py} & Standalone regeneration of empirical vector figures (Figures 7--16) from aggregate CSVs \\
    \code{render_conceptual_diagrams.py} & Mermaid CLI renderer for the six workflow diagrams (\code{diagrams/fig01..fig06*.mmd}) \\
    \bottomrule
  \end{tabular}
  \end{adjustbox}
\end{table}

\bibliographystyle{plainnat}
\bibliography{references}

\end{document}